\documentclass[journal]{IEEEtran}
\usepackage{times}
\usepackage{soul}
\usepackage{url}
\usepackage[hidelinks]{hyperref}
\usepackage[small]{caption}
\usepackage{graphicx}
\usepackage{mathtools}
\usepackage{booktabs}
\usepackage{comment}
\usepackage{tablefootnote}
\usepackage[linesnumbered,ruled,vlined]{algorithm2e}
\SetKwInput{kwInit}{Init}

\SetKwComment{Comment}{$\triangleright$\ }{}
\newcommand{\nonl}{\renewcommand{\nl}{\let\nl\oldnl}}

\usepackage[english]{babel}
\usepackage[utf8x]{inputenc}
\usepackage[T1]{fontenc}
\usepackage{floatrow}
\usepackage{stfloats}

\usepackage{siunitx}
\usepackage{adjustbox}
\usepackage{subcaption}
\usepackage{blindtext}
\usepackage{multirow}
\usepackage{diagbox}

\usepackage{amsmath}
\usepackage{amssymb}
\usepackage{amsfonts}
\usepackage[colorinlistoftodos]{todonotes}
\usepackage{xcolor}
\usepackage{xspace}
\usepackage{color, colortbl}

\usepackage[normalem]{ulem} 
\usepackage{booktabs} 
\usepackage{array}
\usepackage{makecell}

\graphicspath{ {images/} }
\newcommand{\bo}{Bayesian optimization\xspace}
\newcommand{\ea}{evolutionary algorithm\xspace}
\newcommand{\eas}{evolutionary algorithms\xspace}

\ifCLASSINFOpdf
\else
\fi
\begin{document}

%

\title{Time Efficiency in Optimization with a Bayesian-Evolutionary Algorithm}

%
%
%

\author{Gongjin Lan, Jakub M. Tomczak, Diederik M.\  Roijers, A.E. Eiben
\thanks{Gongjin Lan, Jakub M. Tomczak, Diederik M.\  Roijers, A.E. Eiben were with the Department of Computer Science, VU University Amsterdam, Amsterdam, the Netherlands, e-mail: g.lan@vu.nl, j.m.tomczak@vu.nl, diederik.yamamoto-roijers@hu.nl, a.e.eiben@vu.nl}
\thanks{ Diederik M.\  Roijers is with the AI laboratory, Vrije Universiteit Brussel, Belgium; and Institute of ICT, HU University of Applied Sciences Utrecht, the Netherlands.}
}

%
%

\markboth{Journal of \LaTeX\ Class Files,~Vol.
}%
{Shell \MakeLowercase{\textit{et al.}}: Bare Demo of IEEEtran.cls for IEEE Journals}
%



\maketitle

\begin{abstract}
Not all generate-and-test search algorithms are created equal. Bayesian Optimization (BO) invests a lot of computation time to generate the candidate solution that best balances the predicted value and the uncertainty given all previous data, taking increasingly more time as the number of evaluations performed grows. Evolutionary Algorithms (EA) on the other hand rely on search heuristics that typically do not depend on all previous data and can be done in constant time. Both the BO and EA community typically assess their performance as a function of the number of evaluations. However, this is unfair once we start to compare the efficiency of these classes of algorithms, as the overhead times to generate candidate solutions are significantly different. We suggest to measure the efficiency of generate-and-test search algorithms as the expected gain in the objective value per unit of computation time spent. We observe that the preference of an algorithm to be used can change after a number of function evaluations. We therefore propose a new algorithm, a combination of Bayesian optimization and an Evolutionary Algorithm, BEA for short, that starts with BO, then transfers knowledge to an EA, and subsequently runs the EA. We compare the BEA with BO and the EA. The results show that BEA outperforms both BO and the EA in terms of time efficiency, and ultimately leads to better performance on well-known benchmark objective functions with many local optima. Moreover, we test the three algorithms on nine test cases of robot learning problems and here again we find that BEA outperforms the other algorithms.

\end{abstract}


%
\IEEEpeerreviewmaketitle

\section{Introduction}
\label{sec:intr}
When studying and comparing generate-and-test style search algorithms, algorithm efficiency is one of the main factors to consider. To this end, measuring computational effort is essential.
In most existing works, computational effort is measured by the number of iterations or function evaluations, rather than CPU times or wall-clock times. Clearly, this is meant to eliminate effects of particular implementations, software, and hardware, thus making comparisons independent from such practical details \cite{eiben2002a-critical-note}. 
However, this can be very misleading, because the computation time per iteration can differ vastly between different algorithms. This means that comparisons by sample efficiency (measured by the number of iterations) and time efficiency (measured by the computation time) can deliver completely different outcomes. 

\par It is well-known that \bo is the state-of-the-art machine learning framework for optimization tasks in terms of data efficiency, and has been successfully applied in engineering, machine learning, and design \cite{jasper2012practical}. 
However, data efficiency is not the only desirable property of such a generate-and-test search algorithm. 
Particularly, the amount of computation for generating candidate solutions can be prohibitive. 
This is because the computational complexity of BO is $\mathcal{O}(n^3)$ where $n$ is the number of evaluations of candidate solutions, due to the inversion of the co-variance matrix \cite{shahriari2016taking}.\footnote{Although methods have been developed to reduce the computation time \cite{kuindersma2012variational,Gredilla2010sparse,ru2018fast,Snelson2005sparse} (Section \ref{sec:related}), it still grows substantially with the number of evaluations of candidate solutions, limiting its feasibility for longer optimization runs.} 
Alternatively, Evolutionary Algorithms use search heuristics that take constant time for generating candidate solutions, which makes their overhead much less computation-intensive than that of BO, at the expense 
of data efficiency. 
As such, a rule of thumb is that EAs are better for problems where it takes little time to evaluate candidate solutions, and BO is better for the expensive problems, e.g., hyper-parameter tuning. 

\par However, this is not the full story; there is a sizable class of real-world optimization problems that have a moderate evaluation costs \cite{luigi2017practical}. 
It is not clear which algorithms should be chosen then. The key insight behind this paper is that time efficiency and data efficiency should be distinguished and the preference for an algorithm can be based on time efficiency. Taking this perspective, we provide a formal notion of time efficiency by the expected gain in the objective (i.e., fitness) per unit of computation time (Section \ref{sec:efficiency}) and propose to switch from BO to an EA when the time efficiency of BO drops below that of the EA. This results in a new algorithm that 1) runs BO until it becomes `too slow', then 2) selects information obtained by BO and transfers this to an EA by means of a well-balanced initial population, and 3) continues with a special EA that continually monitors the expected gain and uses this information to adjust mutation step-sizes (Section \ref{sec:combining}). 
%
%
We refer to this as Bayesian-Evolutionary Algorithm, BEA for short. 
We compare the performance of BO, the EA, and BEA on a set of well-known objective functions with many local optima (Section \ref{sec:experiment}). 
The experimental results indicate that BEA is not only more time-efficient than BO and EA separately, but also converges to better results.
\par Moreover, we compare the three algorithms on nine test cases in (evolutionary) robotics. To this end, we consider three modular robots with different morphologies, three different tasks, and employ BO, the EA, and BEA to learn well-performing controllers for each combination.\footnote{From the robots' perspective BO, EA, and BEA are learning algorithms that search through the space of controllers. `Under the hood', two of these learning methods, EA and BEA, are evolutionary.}
The experimental results in simulation show that BEA obtains controllers with higher fitness than BO and the EA, while computation times are similar to the EA and much shorter than for BO. Results in simulation are further validated on the physical robots.
%
The experimental results show that the real robots equipped with the best controllers learned by BEA outperform those obtained by BO.

\section{Related work}
\label{sec:related}


\par Evolutionary algorithms are stochastic population-based optimization methods that work by repeatedly applying selection and reproduction operators to a population of candidate solutions \cite{Eiben2015intro}. In general, evolutionary algorithms scale linearly with respect to the problem size and the population size that makes them rather fast optimization algorithms. However, the size of the population plays a crucial role in obtaining good performance. In other words, EAs can suffer from premature convergence. In order to prevent this issue, there were many different techniques proposed, including optimizing EA \cite{golovin2017black}. Here, we propose to initialize the population of the EA by transferring knowledge from a strong optimizer, and also to modify a mutation operator to better adapt to varying improvements and prevent from being stuck in local optima.

\par Currently \bo is the state-of-the-art in generate-and-test search algorithms in terms of sample-efficiency \cite{luigi2017practical,Jones1998efficient}. However, since the computation time is cubic in the total number of function evaluations, \bo can become unfeasible for large numbers of evaluations that greatly limits its applicability. 
In \cite{pourchot2018cem}, Alo\"{ı}s and Olivier proposed a combination scheme using the simple cross-entropy method and Twin Delayed Deep Deterministic policy gradient for evolutionary algorithms to balance the robustness and data efficiency. 
However, the former have the low time efficiency and data efficiency for a large numbers of evaluations, the latter also have the low data efficiency and significantly increase the computational complexity.
\par 
Many studies proposed approaches to reduce the computational complexity using an information-theoretic approach. A recent paper \cite{ru2018fast} developed a fast information-theoretic Bayesian Optimization method, FITBO, that avoids the need for sampling the global minimiser, thus reducing computational complexity. McLeod et al.\ \cite{mcleod2017practical} adapted Predictive Entropy Search and proposed a novel method of selecting sample points to reduce overheads. The other entropy-based methods such as Output-space Predictive Entropy Search \cite{hoffman2015output} and Max-value Entropy Search \cite{wang2017max} improve on Predictive Entropy Search by focusing on the information content in output space. Bai et al.\ \cite{bai2016infor} proposed a method that provides computational efficiency via rapid MAP entropy reduction. 
Other methods aim at reducing the computational effort via sparsification techniques. In \cite{shahriari2016taking} two sparsification techniques for Gaussian processes were reviewed, namely, sparse pseudo-input Gaussian processes \cite{Snelson2005sparse,Seeger2003fastforward} and sparse spectrum Gaussian process \cite{Gredilla2010sparse}. However, all these methods still yield a computation time that grows substantially with the number of evaluations of candidate solutions. Our aim is different: we aim to determine \emph{when} the invested overhead time can no longer be justified by the expected gain, and switch to a method with a much lower computational overhead. In other words, in each iteration we aim to run the most time-efficient alternative from a set of algorithms.

\section{Time Efficiency in Optimization tasks}
\label{sec:efficiency}
A optimization problem is about finding an optimal solution $s \in \mathcal{S}$ of an objective function $f: \mathcal{S} \rightarrow \mathbb{R}$, that is:
\begin{equation}
    s^{*} = \max_{s \in \mathcal{S}} f(s) ,
\end{equation}
where the analytical form of the objective function is usually unknown, i.e., black-box optimization. As a result, calculating gradients is impossible and, thus, a gradient-based optimization algorithm cannot be applied \cite{Jones1998efficient}.

When studying and comparing black-box optimization algorithms, comparisons by \emph{data efficiency} (measured by the number of iterations) and \emph{time efficiency} (measured by  computation time) could lead to completely different conclusions. In this section, we describe a formal framework and the corresponding nomenclature for a discussion of these issues.

Let us consider an iterative optimization algorithm for solving an optimization problem defined by the objective function $f$. 
Without loss of generality, we can assume that $f$ is to be maximized.
Let us assume that the algorithm runs for $\mathcal{N}$ iterations (evaluations), generating $\mathcal{N}$ candidate solutions and recording the time when the candidate solutions are created. Then we use the following notation:
\begin{itemize}
\item $s_i$ is the $i^{th}$ candidate solution, $1 \leq i \leq \mathcal{N}$;
\item $t_i$ is the $i^{th}$ time stamp, $0 \leq i \leq \mathcal{N}$, where $t_0 = 0$ by definition;
\item $f_i = f(s_i) \in \mathbb{R}$ is the $i^{th}$ objective function value, $1 \leq i \leq \mathcal{N}$;
\item $f_i' = \max\{f_i, 1 \leq i \leq \mathcal{N}\}$ is the best objective function value until the $i^{th}$ evaluation.
\end{itemize}

Then, we can define gains (in terms of objective function values), costs (in terms of computation time) and time efficiency (defined as gain per time unit) as follows:  
\begin{itemize}
\item $\delta_i = f_i' - f_{i-1}'$ is the \emph{gain} in the $i^{th}$ iteration, $2 \leq i \leq \mathcal{N}$; 
\item $c_i = t_i - t_{i-1}$ is the \emph{computation time} of the $i^{th}$ iteration, $1 \leq i \leq \mathcal{N}$;
\item $\mathcal{G}_i = \delta_i / c_i$ is the \emph{time efficiency} of the $i^{th}$ iteration, $2 \leq i \leq \mathcal{N}$.
\end{itemize}
\noindent These definitions can be naturally extended for periods of multiple iterations, e.g., 10 iterations, since gain is usually observable for multiple iterations. 
Let $2 \leq i \leq \mathcal{N}$ and $0 < k < i$, $k + 10 = i$ in this paper. Then 
\begin{itemize}
\item $\delta_{(k,i)} = f_i' - f_k'$ is the gain between iteration $k$ and $i$;
\item $c_{(k,i)} = c_i - c_k$ is the computation time between iteration $k$ and $i$;
\item $\mathcal{G}_{(k,i)} = \delta_{(k,i)} / c_{(k,i)}$  is the time efficiency between iteration $k$ and $i$.
\end{itemize}

For generate-and-test style algorithms like BO and EAs, the computation time of one iteration can be naturally divided into the time needed for the \emph{generate} step and the time needed for the \emph{test} step. For instance, in the EA, the \emph{generate} step amounts to performing selection and variation operators to produce a new individual (a candidate solution) and the \emph{test} step amounts to performing a fitness evaluation. Similarly, for BO, an optimization of an acquisition function generates a new candidate that is further tested (evaluated). In general, we can divide the computation time per iteration as well as for a complete run as follows:
$$computation~time = evaluation~time + overhead~time$$
\noindent For a formal discussion, we use the following notions.
\begin{itemize}
    \item $t_{i}^e$ is the $evaluation ~ time$ in iteration $i$,
    \item $t_{i}^o$ is the $overhead ~ time$ in iteration $i$,
    \item $t_{i}^c$ is the $computation ~ time$ in iteration $i$,
\end{itemize}
\noindent where $t_{i}^c = t_{i}^e + t_{i}^o$. Note that $t_{i}^c = c_i$ as they both describe the time needed for one iteration, where $t_{i}^c$ is a general notation and $c_i = t_i - t_{i-1}$ is defined if we have specific experimental data with time stamps $t_i$. 


%


\section{Optimization Algorithms}
\label{sec:bbo_algos}
\par Bayesian optimization and evolutionary algorithms are two types of popular optimization algorithms. 
In this section, we concisely introduce the settings of Bayesian optimization and its cubic computation time, and evolutionary algorithms that the EA in BEA is proposed based on it.

\subsection{Bayesian Optimization}
\label{subsec:bayesian}
\par Bayesian Optimization is a well-known data efficient black-box optimizer which, typically, builds a Gaussian process (GP) approximation of the objective function $f$ of a task. It selects candidate solutions to evaluate on the basis of an acquisition function that balances exploration and exploitation. We present the settings for the GP that we use in this work and why BO takes a cubic increment of computation time. 

\subsubsection{Gaussian processes}
\par A GP is an extension of the multivariate Gaussian distribution to an infinite dimension stochastic process, i.e.,  for any finite set of candidate solutions, $s\in\mathbb{R}^d$, the joint distribution over their objective value, $f(s)\in\mathbb{R}$, is a multi-variate Gaussian distribution \cite{Rasmussen2005gaussian}. Unlike a Gaussian distribution over a random variable, a GP is a distribution over functions, specified by its mean function $m(s)$ and covariance function $k(s_i, s_j)$ (also called as kernel function). This is denoted by $f(x) \sim \mathcal{GP}(m(s), k(s_i,s_j))$. The GP returns the mean and variance of a Gaussian distribution over the possible values of $f$ at $s$. 
A typical choice for the prior mean function is $m(s) = 0$. 
The choice of the kernel, $k(s,s')$, for the GP is crucial, as it determines the shape of the functions that the GP can fit \cite{brochu2010tutorial}. The squared exponential function kernel and Mat\'ern function kernel are popular choices. In this paper, we use the latter. 
 Mat\'ern kernels are parameterized by a smoothness parameter $\nu > 0$. 
 Mat\'ern kernels are a very flexible class of kernels and are commonly used in Bayesian optimization \cite{jasper2012practical,shahriari2016taking,mcleod2017practical}. We use the Mat\'ern 5/2 kernel. This kernel has a hyperparameter $\theta$, the \emph{characteristic length-scale}, which controls the width of the kernel:
\begin{equation}
k_{\nu=5/2}(s,s') = \Big{(} 1 + \frac{\sqrt{5}r}{\theta} + \frac{5r^2}{3\theta^2} \Big{)} \exp \Big{(} -\frac{\sqrt{5}r}{\theta} \Big{)}    
\end{equation}
where $r^2 = (s - s')^T \Lambda (s - s')$, $\Lambda$ is a diagonal matrix of size $d$ with the characteristic length-scale $\theta$ as values on the diagonal. 

\par At iteration $t+1$ of a generate-and-test search algorithm (like BO), we have a dataset, $\mathcal{D}$, containing $s_1, ..., s_t$ previously evaluated solutions, and corresponding values of the objective function $f_1, ..., f_t$ for those solutions (where $f_i=f(s_i)$). The posterior distribution of the objective value for a new -- not yet evaluated -- candidate solution $f(s_{t+1})$, given $\mathcal{D}$, is normally distributed: 
$
\mathcal{P}(f(s_{t+1})| \mathcal{D}) = \mathcal{N}(\mu(s_{t+1}),\sigma^2(s_{t+1}))
$,
with
\begin{align}
    \mu(s_{t+1}|\mathcal{D}) &= k^T\mathbf{K}^{-1}f, \\
    \sigma^2(s_{t+1}|\mathcal{D}) &= k(s_{t+1},s_{t+1}) - k^T\mathbf{K}^{-1}k,
\end{align}
where $k = \begin{bmatrix}
k(s_{t+1},s_1), & k(s_{t+1},s_2), & \dots & k(s_{t+1},s_t)
\end{bmatrix}$, $f = \begin{bmatrix}
f_1, & f_2, & \dots & f_t
\end{bmatrix}$,  and
\begin{equation}
\mathbf{K} = 
\begin{bmatrix}
k(s_1,s_1) & \dots & k(s_1,s_t) \\
\vdots & \ddots & \vdots \\
k(s_t,s_1) & \dots & k(s_t,s_t)
\end{bmatrix}
\end{equation}

\par 
In the derivation process of $\mu(s_{t+1}|\mathcal{D})$ and $\sigma(s_{t+1}|\mathcal{D})$, the computational complexity is dominated by the matrix inversion $\mathbf{K}^{-1}$. 
Standard methods for matrix inversion of positive definite symmetric matrices usually require computational time complexity $\mathcal{O}(n^3)$ for inversion of an $n$ by $n$ matrix \cite{Rasmussen2005gaussian}.
This is why the computation time of Bayesian optimization is cubic in the total number of evaluations.
That is, the time-efficiency of Bayesian optimization decreases for large of numbers of evaluations.
\subsubsection{Acquisition function}
\par To generate new candidate solutions, BO maximizes a so-called \emph{acquisition function} that balances exploitation (i.e., wanting to select candidate solutions with high expected means) and exploration (i.e., the need to evaluate candidate solutions with high uncertainty). 
In this paper, we employ 
\emph{Gaussian Process Upper Confidence Bound} (GP-UCB) that is a state of the art acquisition function \cite{brochu2010tutorial}.
It is defined as follows:
\begin{equation}
  \text{GP-UCB}(s_{t+1}|\mathcal{D}) = \mu (s_{t+1}|\mathcal{D}) + \sqrt{\nu \tau_t} \sigma(s_{t+1}|\mathcal{D}) ,
\end{equation}
where $\nu = 1$, $\tau_t = 2 \log(t^{d/2+2}\pi^2/3\gamma)$ and
$\gamma$ is a constant between 0 and 1 (exclusive), $t$ is the count of evaluations. 
%
\subsection{Evolutionary Algorithms}
\label{subsec:self-adaptive}

Evolutionary Algorithms are population-based, stochastic search algorithms based on the Darwinian principles of evolution \cite{Eiben2015intro}. 
The search process works by iteratively creating a new population that may or may not overlap with the current one. 
The main operators in this loop are (fitness-based) parent selection, reproduction through mutation and/or crossover, and (fitness-based) survivor selection. 
In principle, an EA can work with any type of individuals, including bit-strings, real-valued vectors, permutations, trees, and networks. There are no restrictions on the type of fitness function either.

The test problems we tackle in this paper are real-valued optimization problems with an objective function (fitness function) $f: \mathcal{S} \rightarrow \mathbb{R}$, where $\mathcal{S} \subseteq \mathbb{R}^n$. Hence, a candidate solution in our EA is a vector $(x_1,...,x_j, ..., x_n)$ with $x_j \in \mathbb{R}$ and the reproduction operators need to work on real-valued vectors. 
%
%
%
We use a self-adaptive Gaussian mutation with independent mutation step sizes for each coordinate \cite{Eiben2015intro}. The standard formulas that define the mutated values $x_j'$ for $1 \leq j \leq n$ are as follows 
\begin{align}
    \sigma'_j &= \sigma_j \cdot e^{\tau'\cdot\mathcal{N}(0,1) + \tau \cdot \mathcal{N}_j(0,1)} , \label{eq:mutation_var}\\
    x_j' &= x_j + \sigma'_j \cdot \mathcal{N}_j(0,1) , \label{eq:mutation}
\end{align}
where the $\sigma$ values are the mutation step sizes, $\mathcal{N}(0,1)$ denotes a random number drawn from a Gaussian distribution with 0 mean and standard deviation 1 and $\tau' = 1/\sqrt{2n}$, $\tau = 1/\sqrt{2\sqrt{n}}$ are the so-called learning rates. 
To prevent a standard deviation very close to zero, we use the usual boundary rule with a threshold $\epsilon_0$: $\sigma'_j < \epsilon_0 \Rightarrow \sigma'_j = \epsilon_0$. 
The common base mutation $e^{\tau'\cdot\mathcal{N}(0,1)}$ allows for an overall change of the mutability, while $e^{\tau \cdot \mathcal{N}_j(0,1)}$ provides the flexibility to use different mutation step sizes along different coordinates. This is the basis of the EA in BEA that will be described in (rf. \autoref{subsec:modified_mutation}).

\section{BEA: Bayesian-Evolutionary Algorithm}
\label{sec:combining}

\par The Bayesian-Evolutionary Algorithm (BEA) consists of three stages. In the first stage, i.e., the early iterations, \bo (rf. \autoref{subsec:bayesian}) is employed because its computation time is not yet large. 
The second stage is triggered when the time efficiency, $\mathcal{G}_i$ of the EA becomes greater than that of BO. The role of this stage is to identify and transfer useful knowledge obtained by BO to the EA to be used in the third stage. In the third stage, the search for an optimal solution is continued by an \ea. In general, this can be any EA, but here we introduce and use one with a special mutation mechanism where the mutation step-size is self-adaptive, but is also controlled by the gain. This means that the evolutionary part of BEA uses a hybrid adaptive and self-adaptive mutation strategy to balance exploration and exploitation.
The concise pseudocode of BEA is shown in \autoref{alg:boea}.


\begin{algorithm}[!ht]
	\caption{BEA}
    \label{alg:boea}
    \kwInit{$\mathcal{N}_I$ initial samples $X_{i \in \{1:\mathcal{N}_I\}} (x_1,x_2, ..., x_n)$, $i$ is iteration number; totally iterations $\mathcal{N}$; switch point iteration $\mathcal{N}_s$;}
    \KwResult{solutions $X_{i \in \{1:\mathcal{N}\}} (x_1,x_2, ..., x_n)$ and $f_i$}
    \While{$i <= \mathcal{N}$}
    {
        \uIf {$i < \mathcal{N}_s$}
        {
            run Bayesian optimization       \Comment*[r]{$\mathit{1st\ stage}$}
        }
        \Else{
            \uIf{$i = \mathcal{N}_s$}
            {
                transfer knowledge      \Comment*[r]{$\mathit{2nd\ stage}$}
            }    
                run gain-aware EA         \Comment*[r]{$\mathit{3th\ stage}$}
        }
    }
\end{algorithm}

\subsection{Knowledge Transfer from BO to an EA}
\label{subsec:transfer}

In the second stage of BEA, knowledge obtained by \bo (the first stage) is transferred to the \ea (the third stage). 
In the related literature, knowledge transfer is meant to improve learning on a new task through the transfer of knowledge from a previous task \cite{torrey2010transfer}. 
Our setting is slightly different though; BEA works on a single optimization task and the knowledge is not transferred from one task to another, but between different algorithms on the same task. 
However, the questions to be answered are the same in our context: (1) when to transfer, and (2) what to transfer \cite{Pan2010survey}. 

\par 1) \emph{“When to transfer”} is about picking a moment when the knowledge should be transferred to improve the whole optimization process. As the goal of BEA is to be as time efficient as possible, we propose to employ the gains $\delta_{(k,i)}$ per time-interval, $c_{(k,i)}$, i.e., $\mathcal{G}_{(k,i)}$, to estimate when knowledge should be transferred. Specifically, when the expected value of $\mathcal{G}_{(k,i)}$ becomes less for BO than the EA, BEA should switch from BO to the EA. We refer to this as the \emph{switch point}. 
\par We propose to use the computational efficiency of different algorithms as a metric to determine when should switch from BO to EA. In this work, we calculate the gains, $\delta_{(k,i)}$, of \bo and \eas in the interval from iteration $k$ to $i$, as well as the time it takes to perform these iterations $c_{(k,i)}$. The time efficiency of an algorithm is thus $\mathcal{G}_{(k,i)} = \frac{\delta_{(k,i)}}{c_{(k,i)}}$, as we discussed in Section \ref{sec:efficiency}. Although the time efficiency of both BO and EA usually decrease over the iterations, the time efficiency of BO typically decreases faster than that of EA because of BO's substantially growing overhead time. BEA transfers the useful knowledge from BO to EA when the $\mathcal{G}_{(k,i)}$ of BO becomes less than the $\mathcal{G}_{(k,i)}$ of EA. 
Note that this method of transferring between algorithms can be used for any set of algorithms generally and is not specific to the two algorithms studied in this paper.

\par 2) \emph{“What to transfer”} asks which part of the knowledge that was previously obtained is useful for further learning. 
Specifically, what knowledge obtained by BO can we use in the EA. 
In general, an EA requires solutions with high quality and diversity for balancing exploitation and exploration. We propose four strategies of knowledge transfer that offer a different balance between exploitation and exploration and compare these on the three benchmark objective functions in Section \ref{sec:experiment}. 

\par Previous work on knowledge transfer mostly focuses on what kind of knowledge is useful for a task. For example, a straightforward type of knowledge transfer is to provide the initial solutions for a new task on the basis of a previous task. This is called a \emph{starting-point method} \cite{torrey2010transfer}. We use a conceptually similar approach to kick-start the \ea in BEA.

\par In \eas, population initialization affects the convergence speed and also the quality of the final solution \cite{Rahnamayan2007novel}. Specifically, the quality (which can be exploited) and the diversity (which assures appropriate exploration) of the initial solutions are crucial. We thus aim to transfer a high-quality and diverse set of solutions from BO to the EA. We denote the population size of the \ea in BEA as $p$, the set of all solutions from \bo by $\mathcal{S}_{BO}$ and propose the following strategies for knowledge transfer:
\begin{itemize}
    \item \textbf{S1}: \textit{The last p solutions} in $\mathcal{S}_{BO}$ are transferred to the EA. This is the simplest method. While the last $p$ candidate solutions visited by BO need not be diverse, their quality is typically good.
    \item \textbf{S2}: \textit{The top $p$ solutions} in $\mathcal{S}_{BO}$ are transferred to the EA. The selected solutions are the best in terms of objective value, guaranteeing the highest possible quality BO can offer. However, the diversity of this set is not guaranteed. 
    \item \textbf{S3}: \textit{K-means clustering}, where $K = p$. All candidate solutions in $\mathcal{S}_{BO}$ are divided into $p$ clusters. The best solution in each cluster is transferred to the EA. This strategy tries to provide diversity and quality. However, some of the selected solutions can be of low quality if their cluster consists of poor solutions. 
    \item \textbf{S4}: \textit{K-means clustering, where $K = p$, in the top $d\%$ solutions}. To make a good trade-off between quality and diversity we first select a set of good solutions, then enforce diversity by clustering and choose the best solution from each cluster. 
    In this work, we set $d = 50$, i.e., use half of the solutions generated by BO. 
\end{itemize}

\subsection{A Gain-aware Evolutionary Algorithm}
\label{subsec:modified_mutation}
In principle, any EA can be used in the third stage of BEA, but here we employ one with a new mutation operator that makes use of the available gain information $\delta_i$. The idea is to use mutation with self-adaptive step-sizes based on modifying \autoref{eq:mutation}, as
\begin{equation}
x_j' = x_j + \sigma'_j \cdot \mathcal{N}_j(0,1) \cdot \sigma''_i
\end{equation}
where $j$ denotes the coordinates of the objective function ($1 \leq j \leq n$, $n$ is dimension of a solution, rf. \autoref{subsec:self-adaptive}), $i$ is the iteration counter ($1 \leq i \leq \mathcal{N}$) and the value of $\sigma''_i$ is adapted throughout the evolutionary run as follows. We set $\sigma''_0 = 1$ and update its value at each iteration by
\begin{equation}
 \sigma''_i = \begin{cases}
    \alpha \cdot \sigma''_{i-1} \quad \quad \text{if}~ \delta_{(k,i)} = 0\\
    \beta \cdot \sigma''_{i-1} \quad \quad \text{if}~ \delta_{(k,i)} \neq 0 
 \end{cases}
\end{equation}
\noindent where $\alpha > 1$, $\beta < 1$ are parameters, $\delta_{(k,i)}$ is the gain between iteration $k$ and $i$ as we discussed in Section \ref{sec:efficiency}.

 Note, that using different $\sigma''$ values it is possible to do more exploitation (smaller mutation step sizes) or more exploration (larger mutation step sizes). We will use this possibility to fight premature convergence and increase the extent of exploration if the gains are diminishing. The above method produces smaller mutation step size for exploitation to efficiently search better solutions when algorithm has no gains during long periods of iterations and larger mutation step sizes for exploration when the algorithm has no gains during long periods of iterations. Empirically, $\alpha$ and $\beta$ are set to close 1 that not too large or small, which prevent sharp fluctuations of mutation step sizes. 

\par A further change to the standard mutation operator concerns the way we handle cases at the boundaries of the search space. Usually, the parameter $x_j$ in a solution $(x_1,...,x_n)$ is forced to the boundary value $\mathcal{L}_j$ or $\mathcal{U}_j$ when mutation would place it outside its domain $(\mathcal{L}_j,\mathcal{U}_j)$. Consequently, solutions with values that are exactly on the boundary are often evaluated repeatedly, which is undesirable. To mitigate this effect, we propose to repeat Gaussian mutations until the mutated value $x'_j$ is in the domain $(\mathcal{L}_j,\mathcal{U}_j)$.
The pseudocode of the modified self-adaptive Gaussian mutation about generating a new solution, is shown in Algorithm \ref{alg:modified_mutation}.
\begin{algorithm}[!ht]
	\caption{Gain-aware self-adaptive Gaussian mutation}
    \label{alg:modified_mutation}
	\KwIn{A solution $(x_1,x_2, ..., x_j, ..., x_n)$, $\sigma''_i$, $\delta_{(k,i)}$}
    \KwOut{A new solution $(x'_1, x'_2, ..., x_j, ..., x'_n)$}
        $\tau \leftarrow 1/\sqrt{2n}$, $\tau' \leftarrow 1/\sqrt{2\sqrt{n}}$\\
        $\sigma'_j \leftarrow \sigma_j \cdot e^{\tau'\cdot\mathcal{N}(0,1) + \tau \cdot \mathcal{N}_j(0,1)}$\\
        \uIf {$\delta_{(k,i)} = 0$}
        {
        $\sigma''_j  \leftarrow \alpha \cdot \sigma_{j-1}''$       \Comment*[r]{$\mathit{more\ exploration}$}
        }
        \Else{
        $\sigma''_j  \leftarrow \beta \cdot \sigma_{j-1}''$      \Comment*[r]{$\mathit{more\ exploitation}$}
        }
        $x'_j \leftarrow x_j + \sigma'_j \cdot \sigma''_i \cdot (\text{a\ sample\ from\ }\mathcal{N}_j(0,1) )$ \\
        \While{$x'_j < \mathcal{L}_j$}
        {$x'_j \leftarrow$ sample from $\mathcal{N}(\mathcal{L}_j,\sigma'_j)$}
        \While{$x'_j > \mathcal{U}_j$}
        {$x'_j \leftarrow $ sample from $\mathcal{N}(\mathcal{U}_j, \sigma'_j)$}
        \Return{$(x'_1, x'_2, ..., x'_j, ..., x'_n)$}
\end{algorithm}

\section{Experiments on Synthetic Test Functions}
\label{sec:experiment}
\subsection{Setup}
\label{subsec:setup}
\par To test the performance of BO, the EA and BEA, we measure their performance on three well-known benchmark objective functions from the evolutionary computation community \cite{Jamil2013literature,Molga2005test,pohlheim2007examples}: Schwefel, Griewank, and Rastrigin with $d = 20$ dimensions. For each of these functions, the global optimum is zero located at the origin. 

In all experiments, we use standard Bayesian optimization from the flexible high-performance library Limbo \cite{cully2018limbo}, using a Gaussian process with a Mat\'ern 5/2 kernel and the length scales ($\theta = 0.1$ for Griewank and Rastrigin, $\theta = 0.5$ for Schwefel), and a GP-UCB acquisition function (as specified in Section \ref{subsec:bayesian}), for both the BO and BEA. This hyperparameter setting outperforms other hyperparameter settings in our preliminary experiments on hyperparameter tuning.
For the constituent components of BEA, we run the exact same BO with the same hyperparameter setting in the first stage of BEA.


\par As for the evolutionary algorithm, we use a real-valued encoding with arithmetic recombination, self-adaptive Gaussian mutation, tournament parent selection and elitist survivor selection for both the standalone EA and the EA in BEA. 
The difference is that we propose a gain-aware self-adaptive Gaussian mutation (Algorithm \ref{alg:boea}) for the EA in BEA to adjust exploration and exploitation based on the gain. 
In this paper, we use the hyperparameter values $\alpha = 1.03$, $\beta = 0.99$ for the gain-aware self-adaptive Gaussian mutation.
For parameters of both EA and BEA, the mutation rates are 0.8, the population sizes are 10, the tournament sizes are 2.
In the preliminary hyperparameter tuning, we selected the crossover rate value of $0.7$ for the EA and the lower crossover rate value of $0.1$ for the EA in BEA to reduce the exploration and encourage the exploitation for further learning.
\subsection{Number of Evaluations vs. Computation Time} 
\label{subsec:results}

\par As discussed in Section \ref{sec:efficiency}, the computation time at the $i^{th}$ iteration has two components: the evaluation time $t^e_i$, and the overhead time $t^o_i$ and algorithms can differ much in their overhead times. The net effect of more overhead --and thus the preference for a given algorithm-- obviously depends on the ratio between evaluation times and overhead times. To study these effects systematically, we should be able to vary the evaluation times to simulate different scenarios. To this end, we introduce $t^e_i$ as a parameter and assume that the evaluation times generally do not vary per iteration $i$, nor by the evaluated candidate solutions $s_i$ for a given task.
By these assumptions, we can use $t^e$ to denote the evaluation time and make plots of algorithm behavior over computation time for different values of $t^e$. Given an objective function and the usual plot of objective function values over evaluations (based on actual measurements by running the given algorithm), a curve that shows objective function values over computation time can be made for any value of $t^e$. The data points in this plot are calculated by using $t_i = \sum^i_1 (t^o_i + t^e)$, i.e., the total computation time for all $i$ iterations. 

\par
\begin{figure}[!htbp]
$\begin{array}{rl}
    \hspace{-0.15cm} \includegraphics[width=0.49\textwidth]{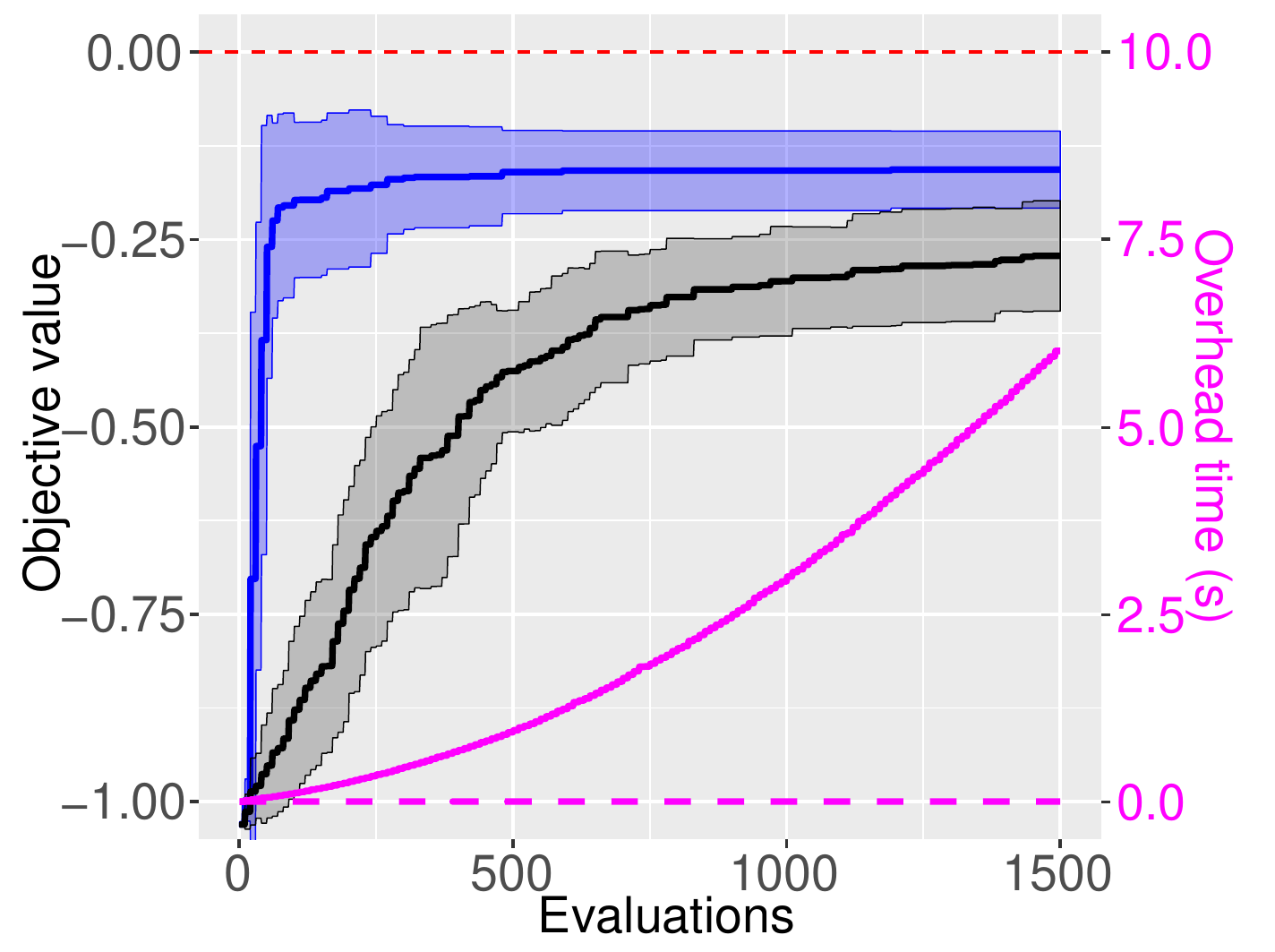} & 
    \hspace{-0.3cm} \includegraphics[width=0.5\textwidth]{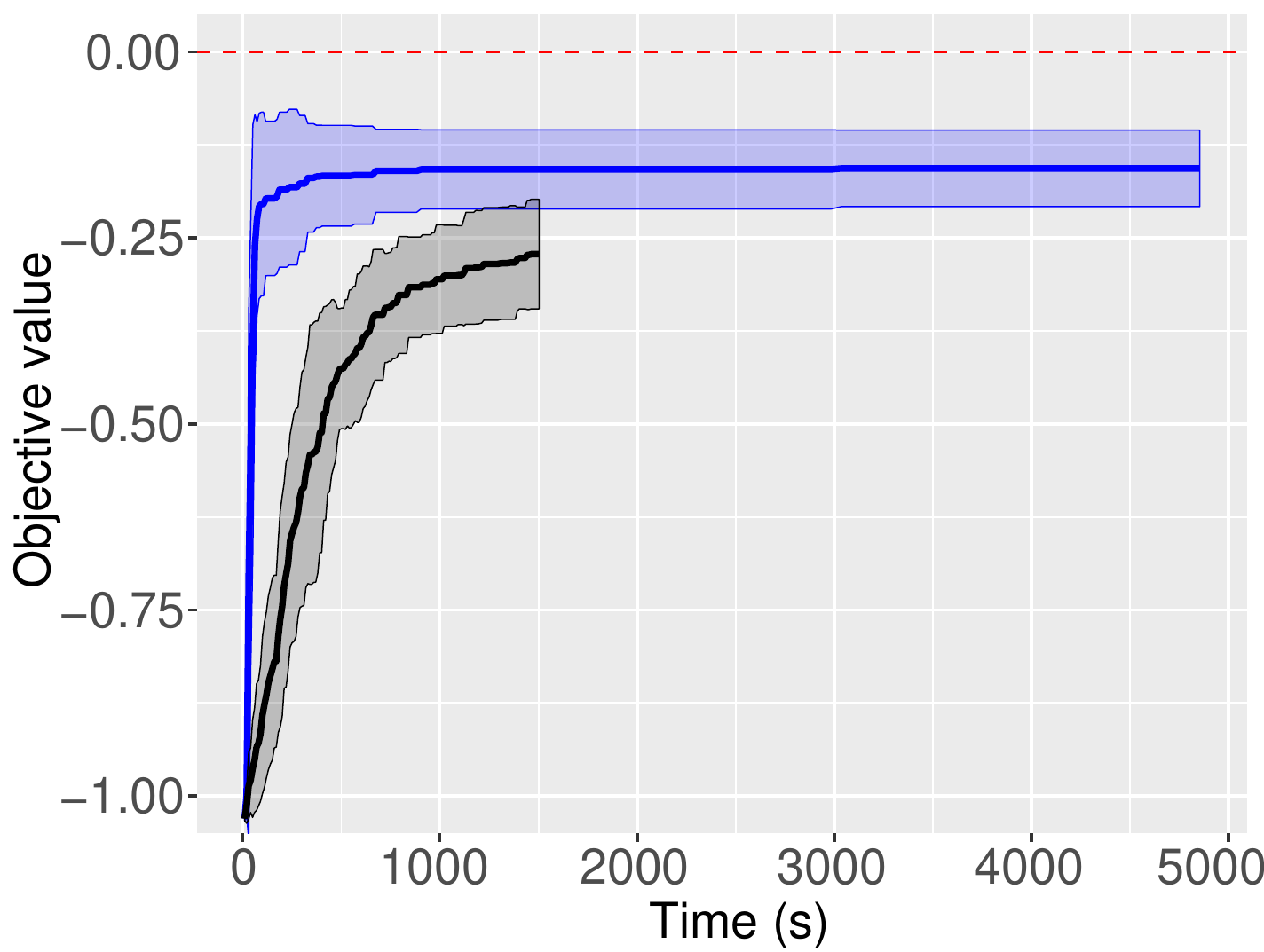}\\
    \multicolumn{2}{c}{\includegraphics[width=0.25\textwidth]{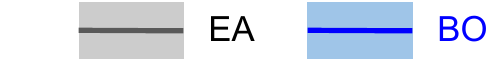}}
\end{array}$
\caption{The objective value and overhead time of BO and EA on the Griewank function. Left: the blue and black show the objective value over the number of evaluations for BO and EA respectively; the magenta solid and dashed lines represent the overhead time of BO and EA respectively. 
Right: the blue and black show the objective value over computation time (the evaluation time was set to 1 second) for BO and EA respectively. The shadows are the 95.45\% confidence areas (two standard deviations). }
\label{fig:griewank}
\end{figure}

 To gain insight into the difference in perspective when it comes to objective values as a function of the number of evaluations performed compared to as a function of computation time, we run BO and the EA on one of the objective functions, the Griewank function, to obtain the objective values and overhead times. Then we calculate the computations times for $t^e=1s$ and plot the objective values over computation time as described above. The results are shown in \autoref{fig:griewank}.

\par First, we observe that BO obtains a better objective value within a smaller number of evaluations.
However, when we take the perspective of objective value as a function of computation time (i.e., including the evaluation time), EA uses only a fraction of the time compared to BO. 
BO takes a rapid increment of overhead time. In contrast, the overhead time of EA stay at a constant level near to zero.
Furthermore, at the end of its curve, EA obtains higher gain with less computation time, i.e., higher gain per second ($\mathcal{G}_i$), which means that the issue is still inconclusive as to which algorithm is better.
In short, one might mistakenly think that an algorithm definitely outperforms another when looking at the objective value as a function of the number of evaluations performed, rather than as a function of computation time. 

\subsection{Gain and Switch Point}
\begin{figure*}[!htbp]
    \centering
    \begin{adjustbox}{max width=0.95\columnwidth}
    \begin{tabular}{c c c}
        Evaluation time: 0.1 second & Evaluation time: 1 seconds & Evaluation time: 10 seconds \\
        \includegraphics[width=0.49\textwidth]{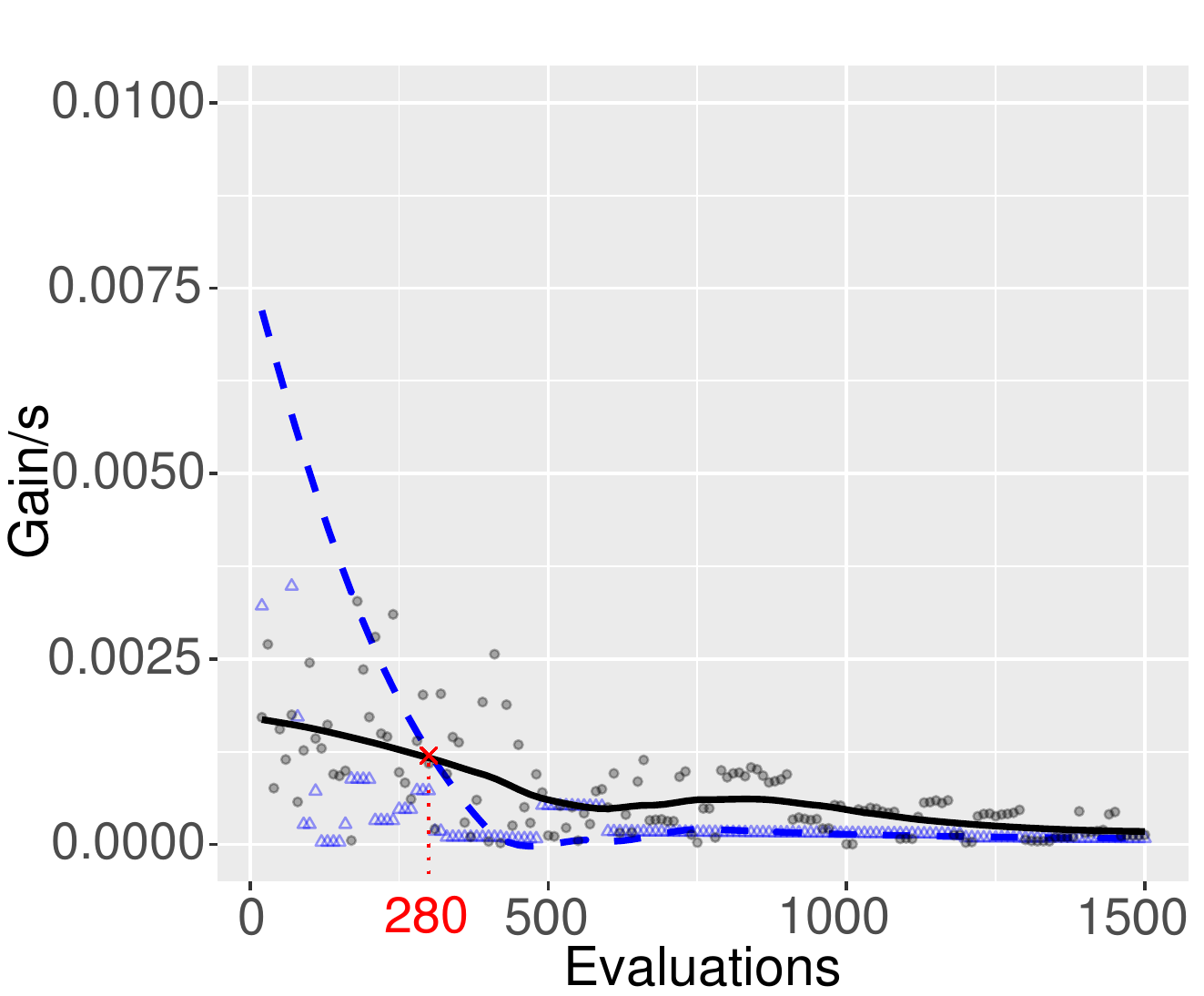}&
        \includegraphics[width=0.49\textwidth]{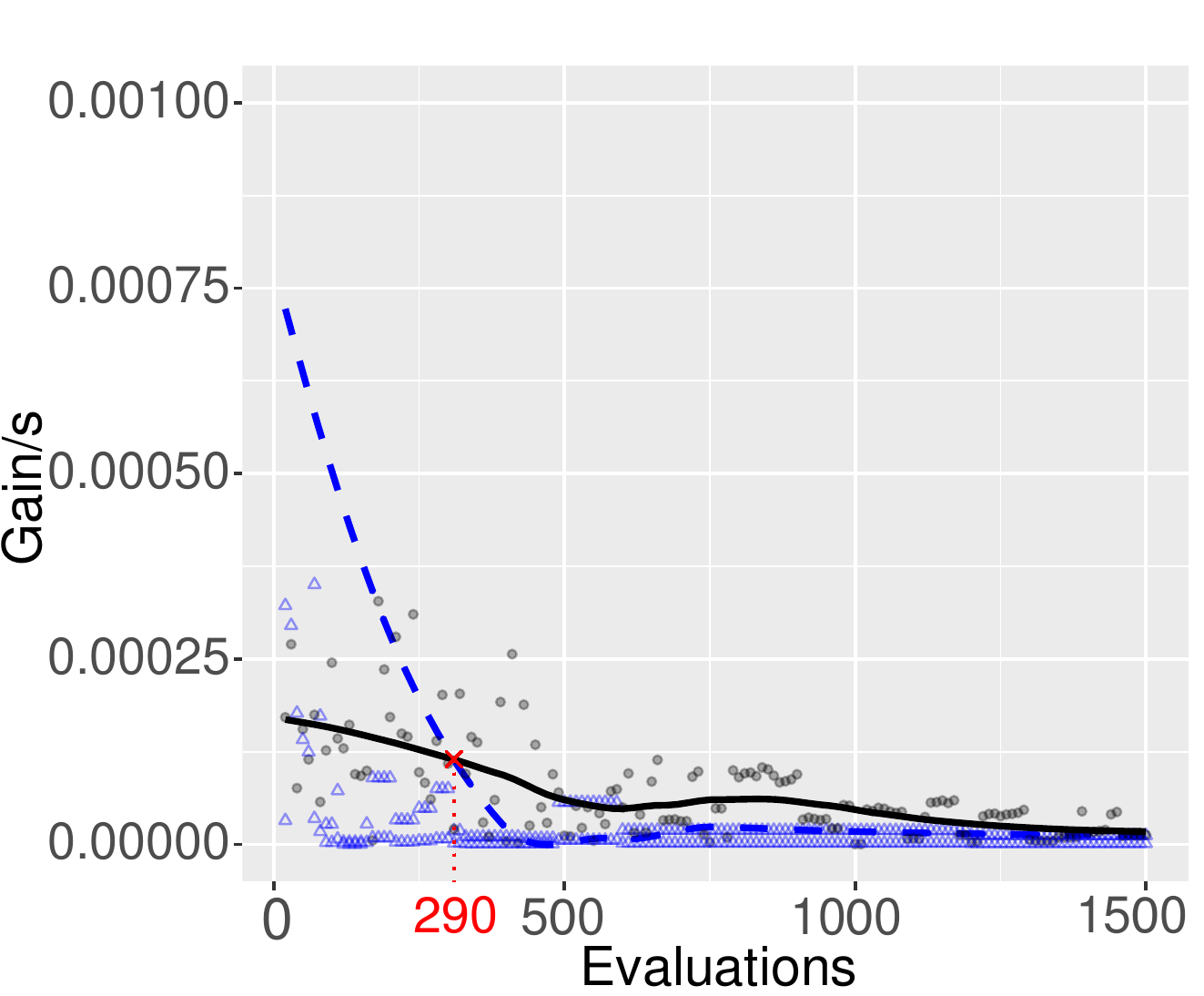}&
        \includegraphics[width=0.49\textwidth]{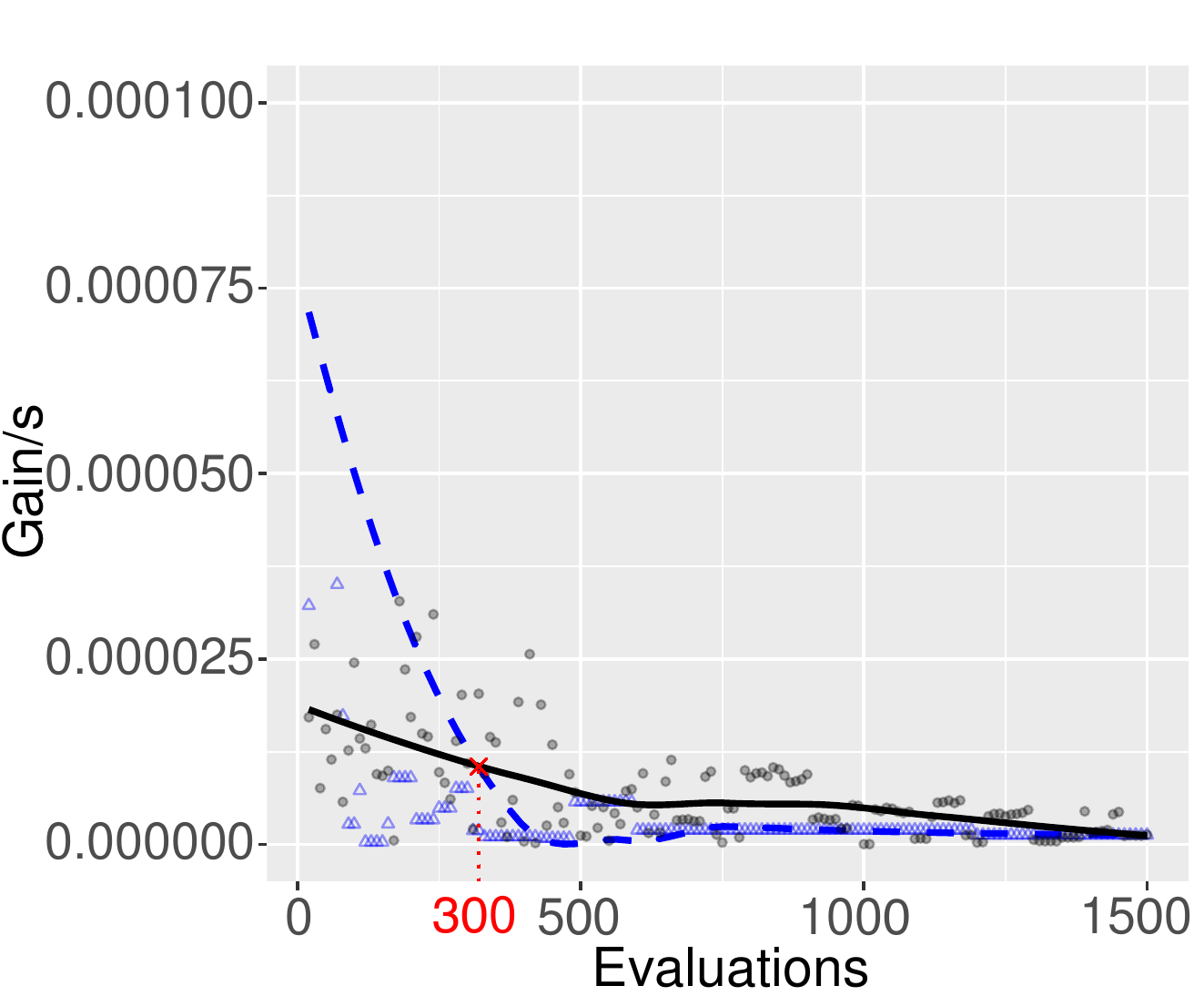}\\
            & Griewank function & \\
        \includegraphics[width=0.49\textwidth]{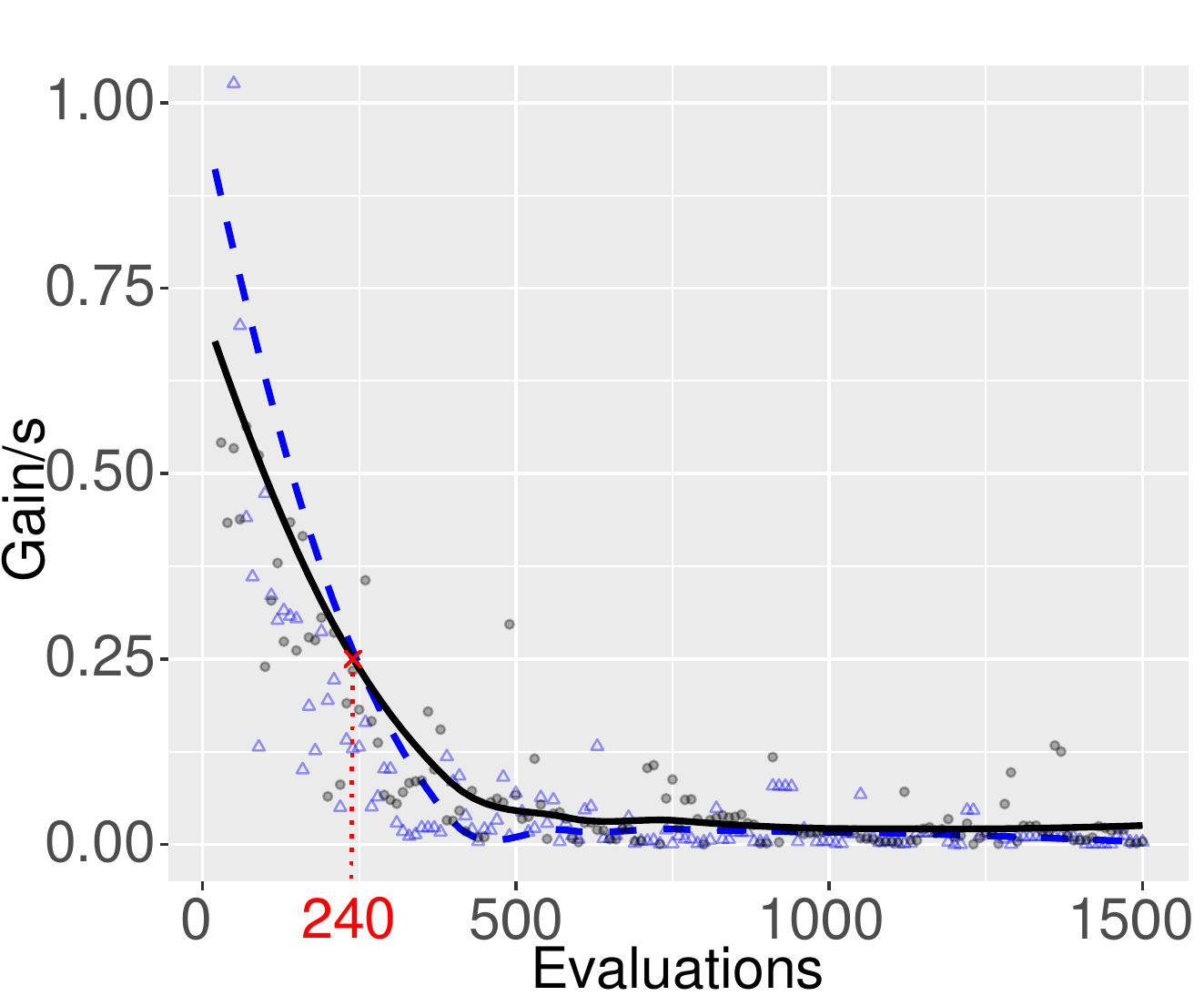}&
        \includegraphics[width=0.49\textwidth]{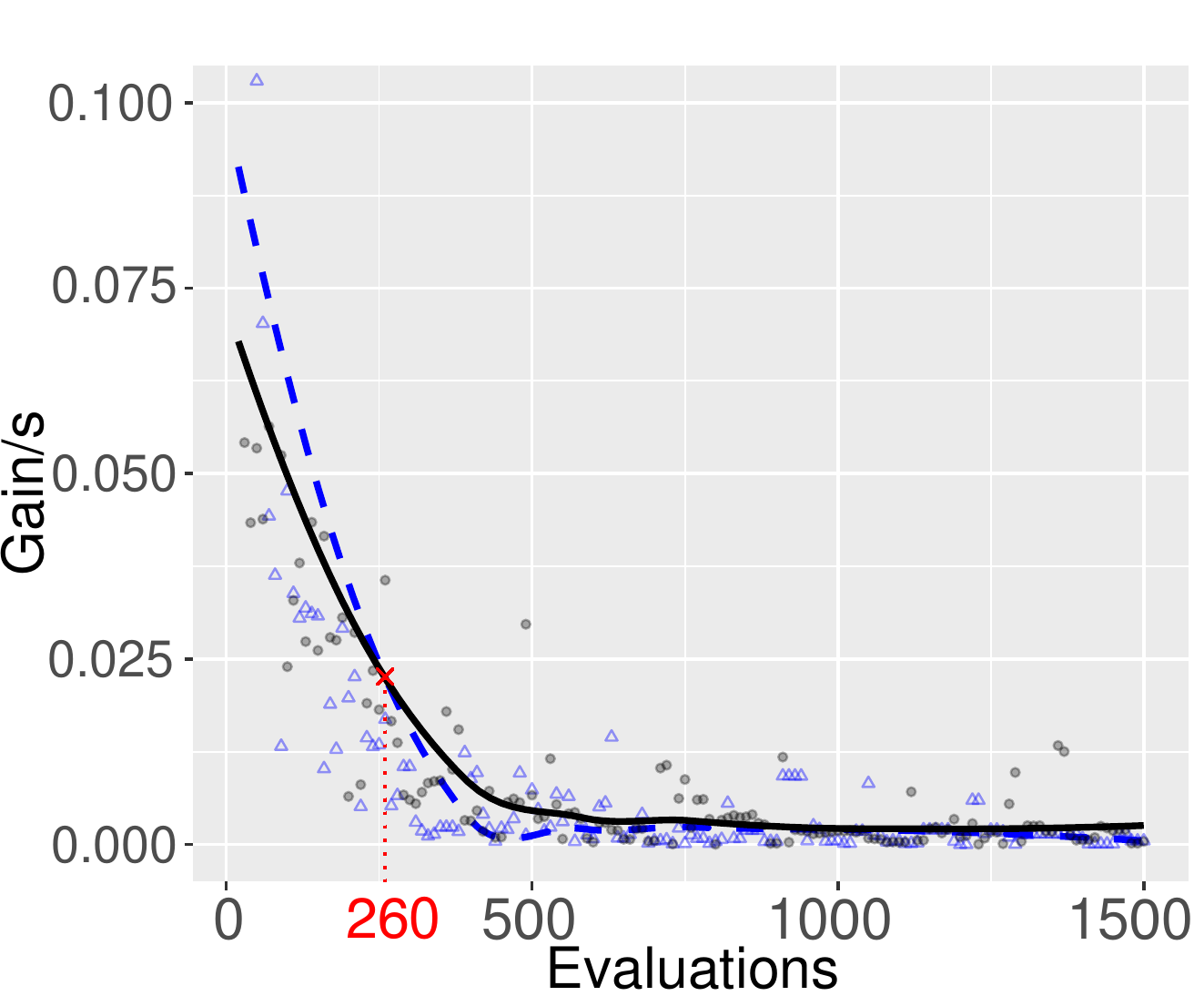}&
        \includegraphics[width=0.49\textwidth]{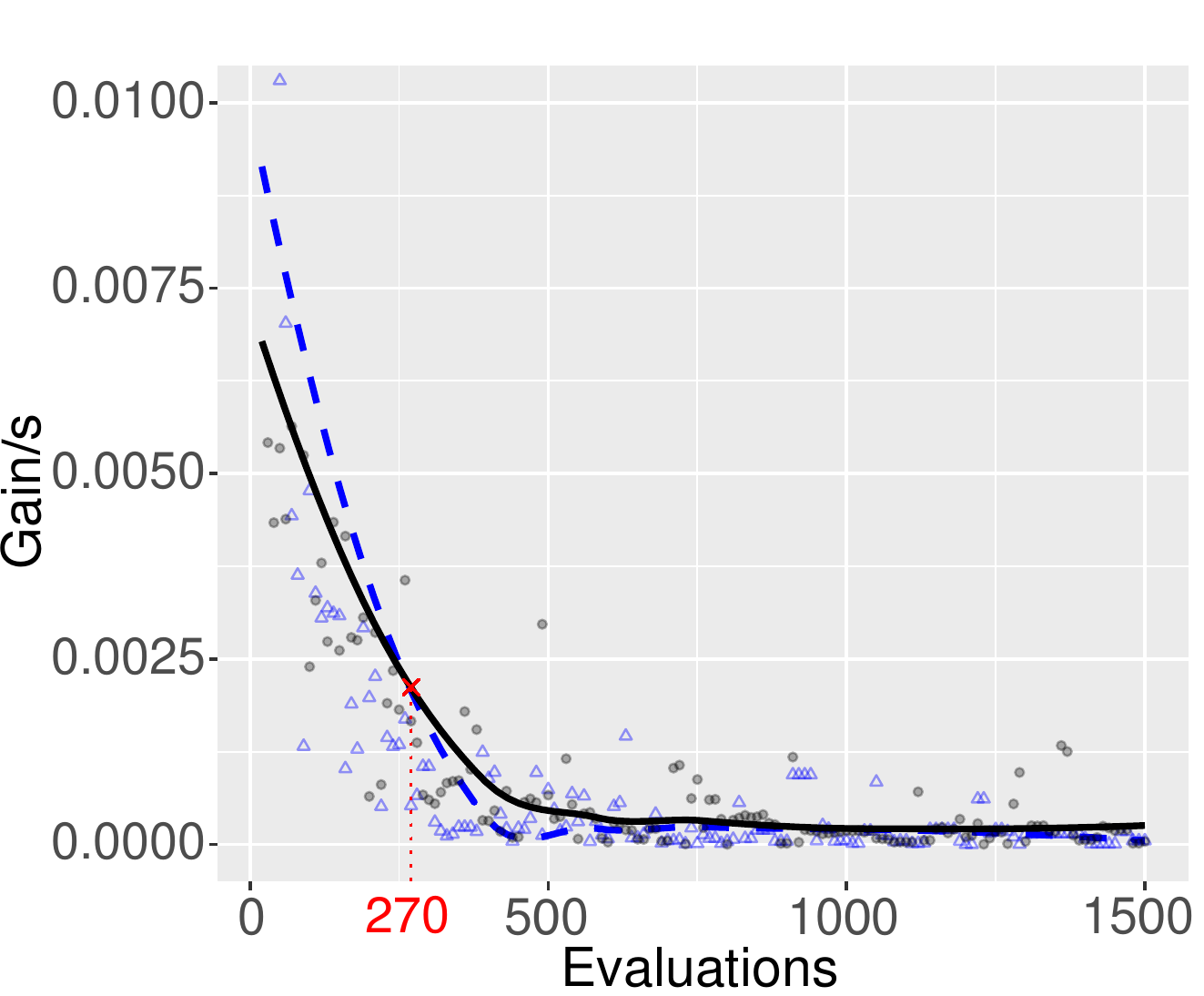}\\
            & Rastrigin function & \\
        \includegraphics[width=0.49\textwidth]{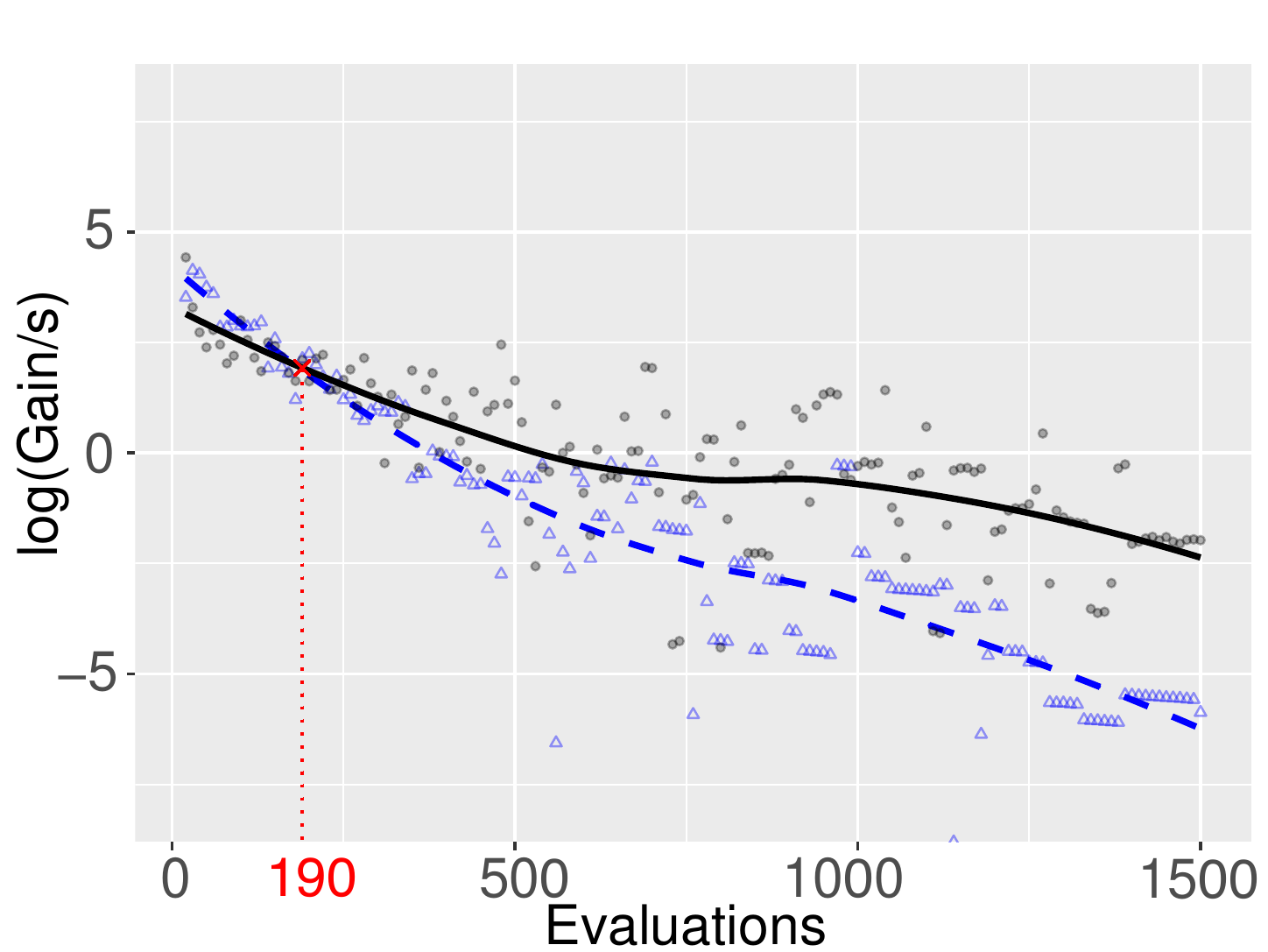}&
        \includegraphics[width=0.49\textwidth]{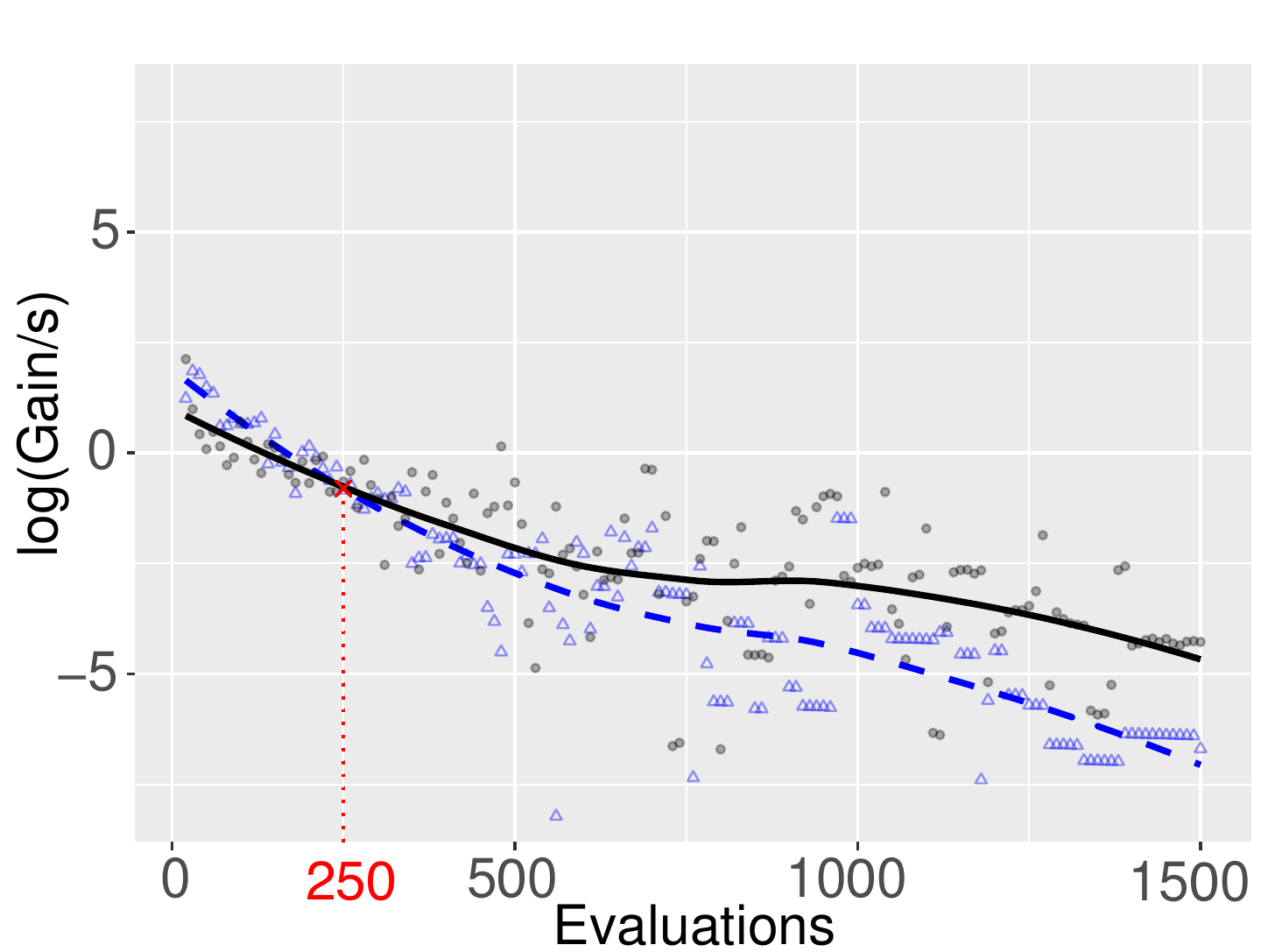}&
        \includegraphics[width=0.49\textwidth]{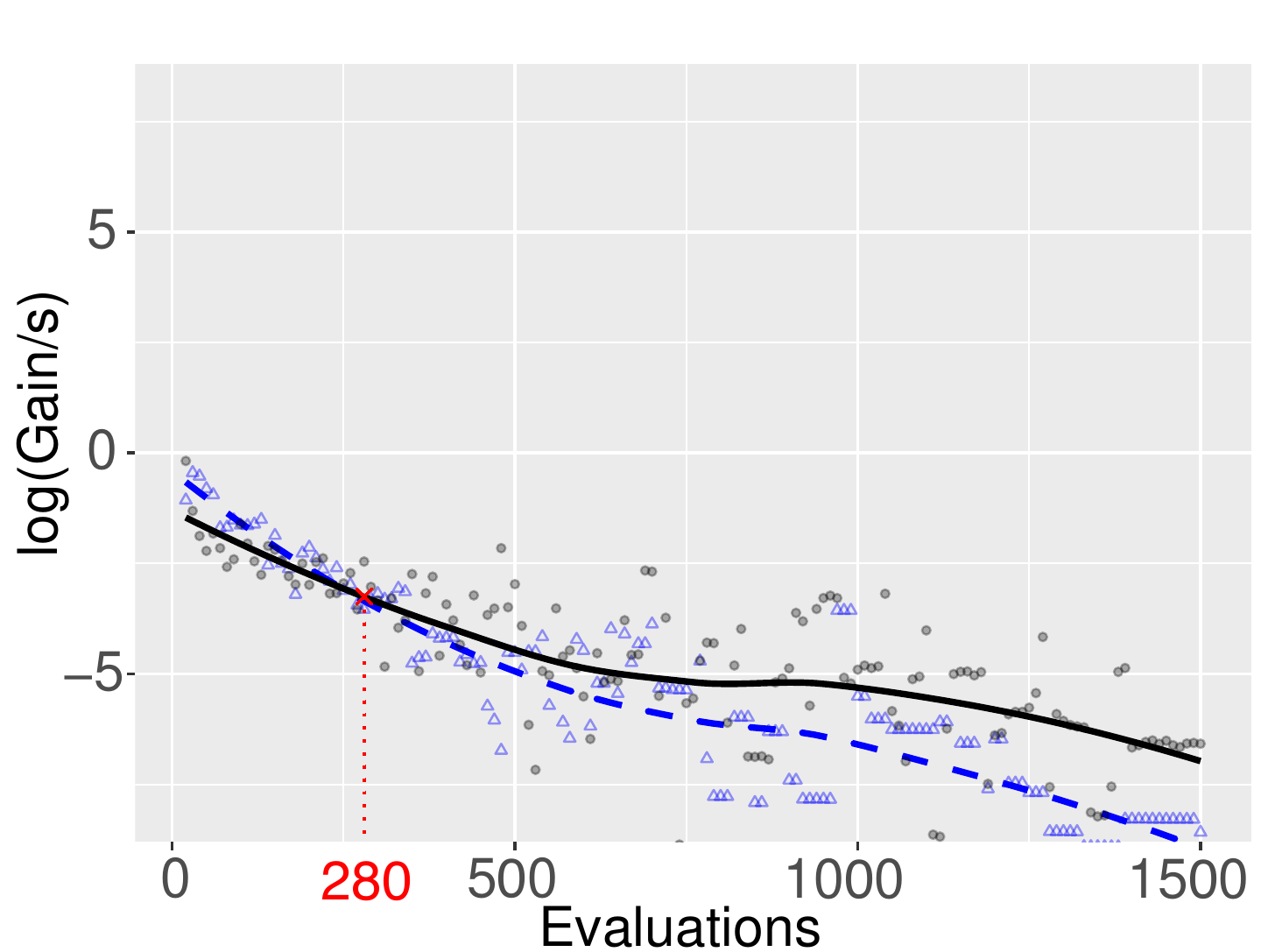}\\
            & Schwefel function & \\
            & \includegraphics[width=0.25\textwidth]{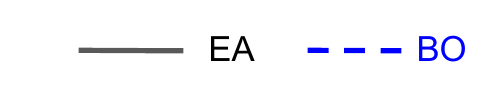} & \\
    \end{tabular}
    \end{adjustbox}
    \caption{Gain/s, log(Gain/s) of Bayesian optimization (blue) and the evolutionary algorithm (black) with different evaluation times, 0.1s, 1s, 10s. The red dashed lines and dots present the intersections (switch points) where Bayesian optimization and evolutionary algorithm have the same value of Gain/s, log(Gain/s). Note that Gain/s is the time efficiency $\mathcal{G}_{(k,i)}$.}
    \label{fig:gain/s-bo-ea}
\end{figure*}

Our perspective is that an algorithm is computationally efficient if it has a high expected gain in objective value per unit of computation time (Section \ref{sec:efficiency}). 
To compare BO and the EA, we measure their gain per second averaged over ten evaluations, for the three benchmark objective functions and the situations of three values of evaluation time, $t^e \in \{0.1s, 1s, 10s\}$. The results are presented in \autoref{fig:gain/s-bo-ea}.
\par For these values of $t^e$, BO is more computationally efficient in its early iterations, while EA becomes more computationally efficient in the later iterations. We denoted the point where EA overtakes BO in terms of the computational efficiency by a red cross. 
We observe that as $t^e$ increases, the iteration number at which EA overtakes EA becomes larger as well but this is a relatively small increase. The largest value for $t^e$ is hundred times the smallest value, but the takeover point only increases about $10\% - 40\%$.    
Furthermore, we notice that while this point differs for various objective functions, it lies in the interval between $190$ and $300$ iterations. 
On the basis of this experiment, we suggest $250-300$ iterations as a heuristic value for the switch point in BEA. In the sequel, we use $250$ for BEA on the numerical objective functions and $300$ on the robot learning test cases. 

\subsection{Strategies for Knowledge Transfer}

\begin{figure*}[!htbp]
    \centering
    \begin{adjustbox}{max width=0.95\columnwidth}
    \begin{tabular}{c c c}
        \includegraphics[width=0.49\textwidth]{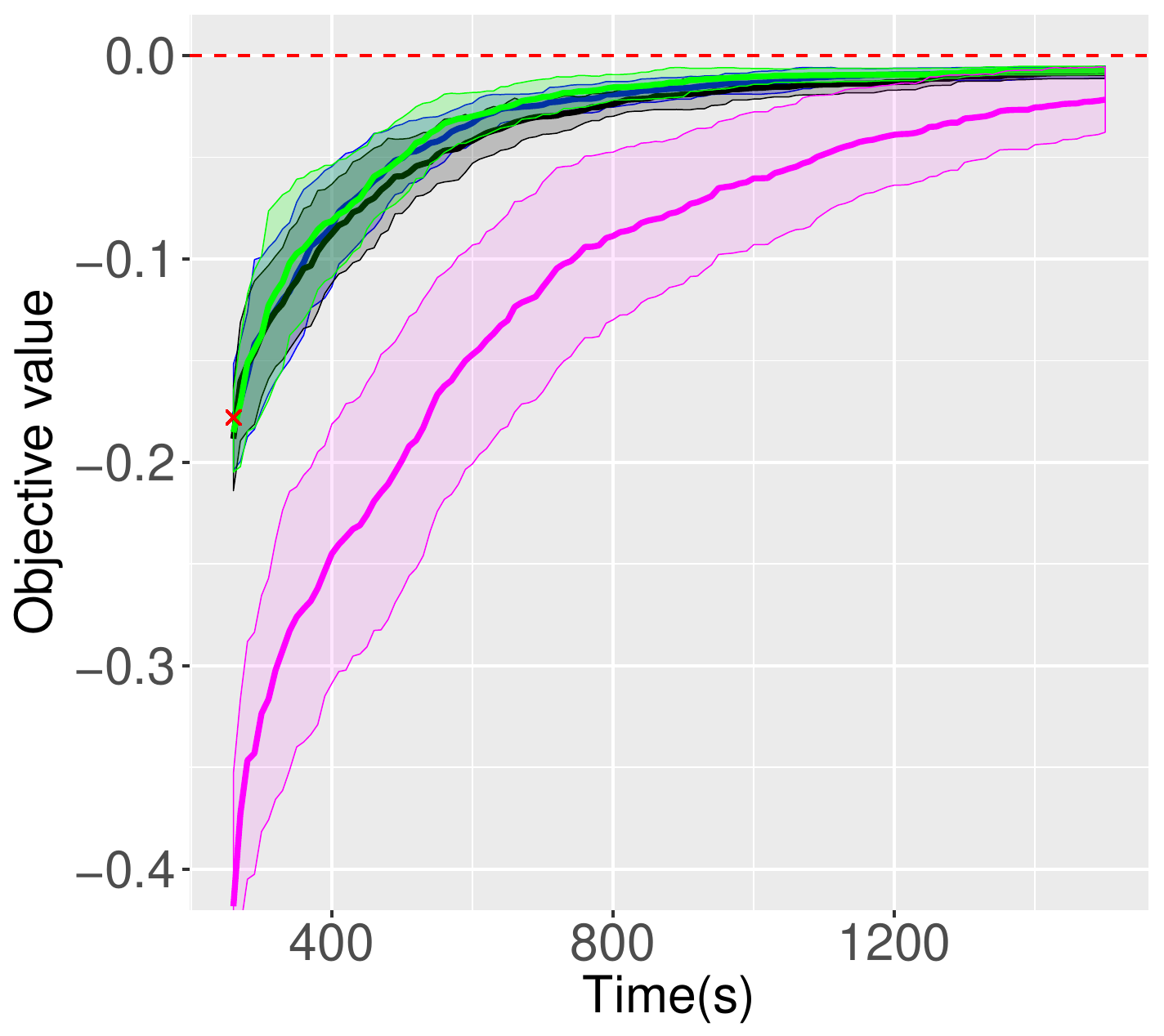}&
        \includegraphics[width=0.49\textwidth]{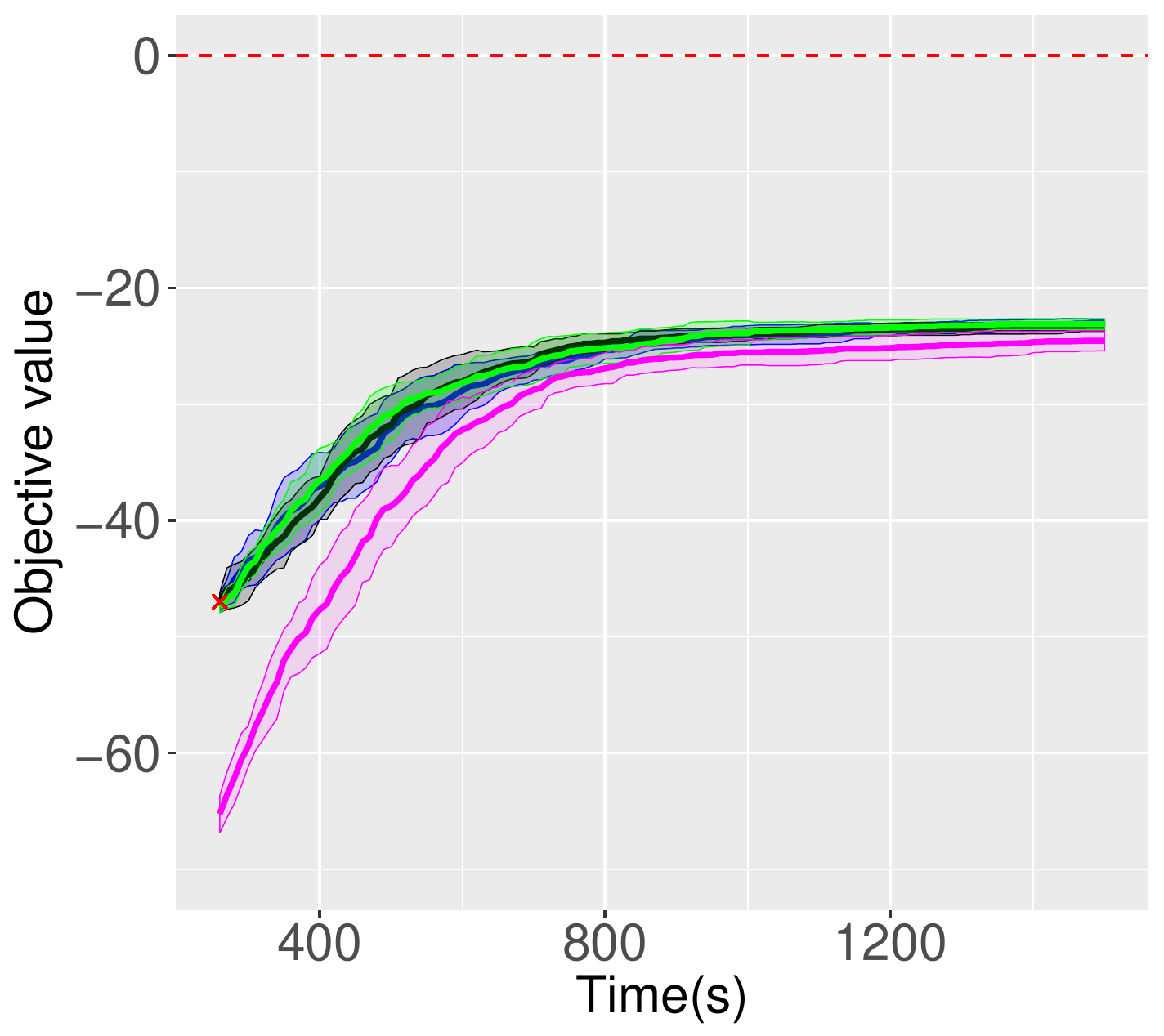}&
        \includegraphics[width=0.49\textwidth]{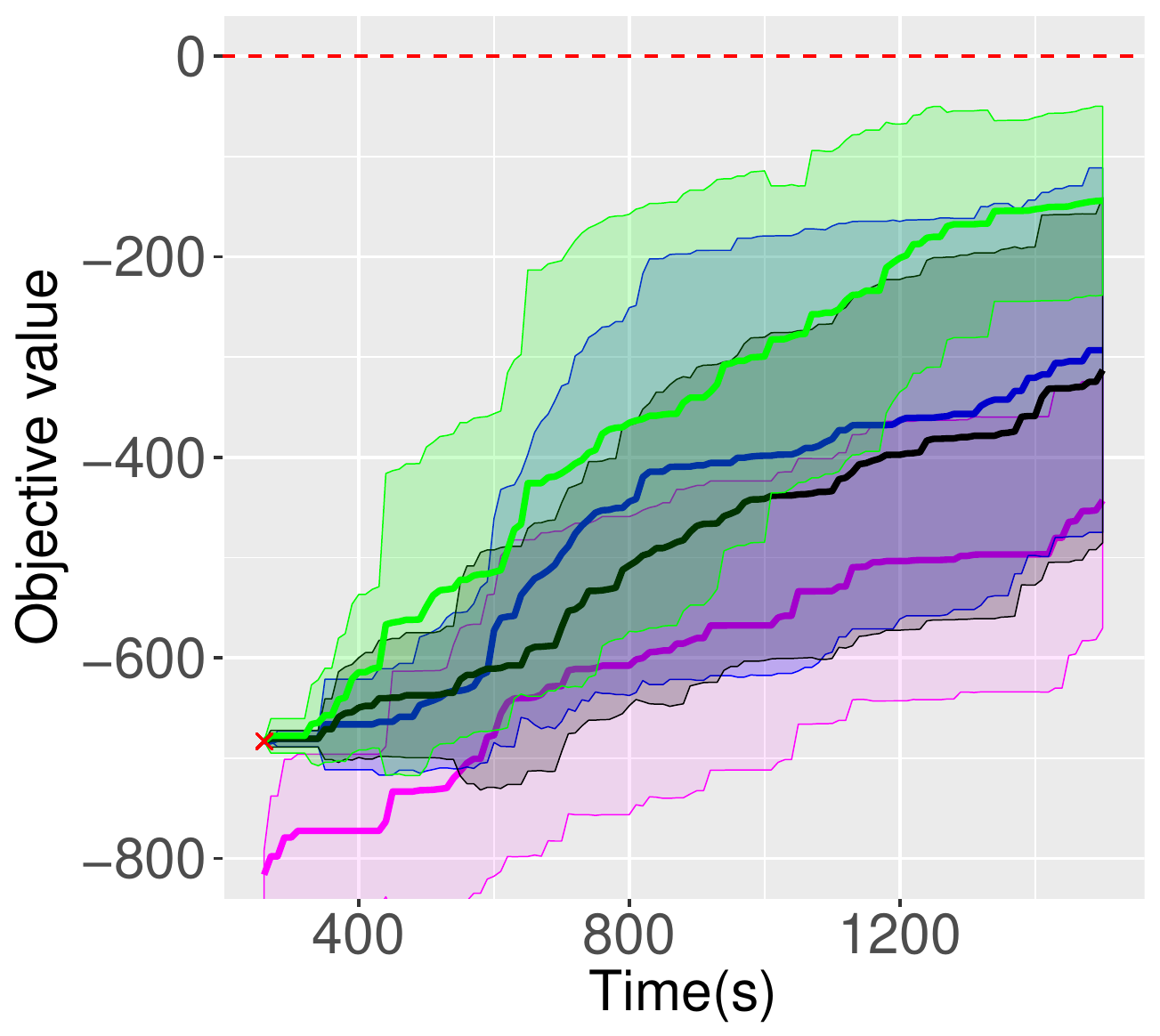}\\
        Griewank function & Rastrigin function & Schwefel function \\
            & \includegraphics[width=0.3\textwidth]{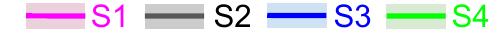} & \\
    \end{tabular}
    \end{adjustbox}
    \caption{The objective value as a function of computation time (evaluation time $t^e=1 s$) for BEA using the strategies S1 (magenta), S2 (blue), S3 (black), and S4 (green) when switching from BO to the EA. The center lines are the averages of 20 repetitions. The shadows show the 95.45\% confidence area (two standard deviations). 
    The dashed lines are optimal values of the three test functions.
    }
    \label{fig:transfer}
\end{figure*}

In Section \ref{subsec:transfer}, we described four possible strategies of knowledge transfer, S1, S2, S3, and S4. To compare these strategies, we test them on all three objective functions and inspect the growth curves of the EA in BEA after the knowledge transfer, as shown in \autoref{fig:transfer}. 

From the curves in \autoref{fig:transfer}, we notice that S1 is dominated by the other transfer methods. The differences between the other three approaches are insignificant for the Griewank and Rastrigin objective functions (though S4 has marginally higher values). For the Schwefel function, S4 outperforms S2 and S3 clearly. The difference between S4 and S2 and S3 are significant (with p-values of $<0.02$ and $<0.04$, respectively). We therefore decide to use S4 as the strategy of knowledge transfer in BEA to further compare the EA, BO, and BEA.
%
\subsection{Evaluation over Computation Time}
\begin{figure*}[!htbp]
    \centering
    \large
    \begin{adjustbox}{max width=1.0\columnwidth}
    \begin{tabular}{c c c}
        0.1 second & 1 seconds & 10 seconds \\
        \includegraphics[width=0.49\textwidth]{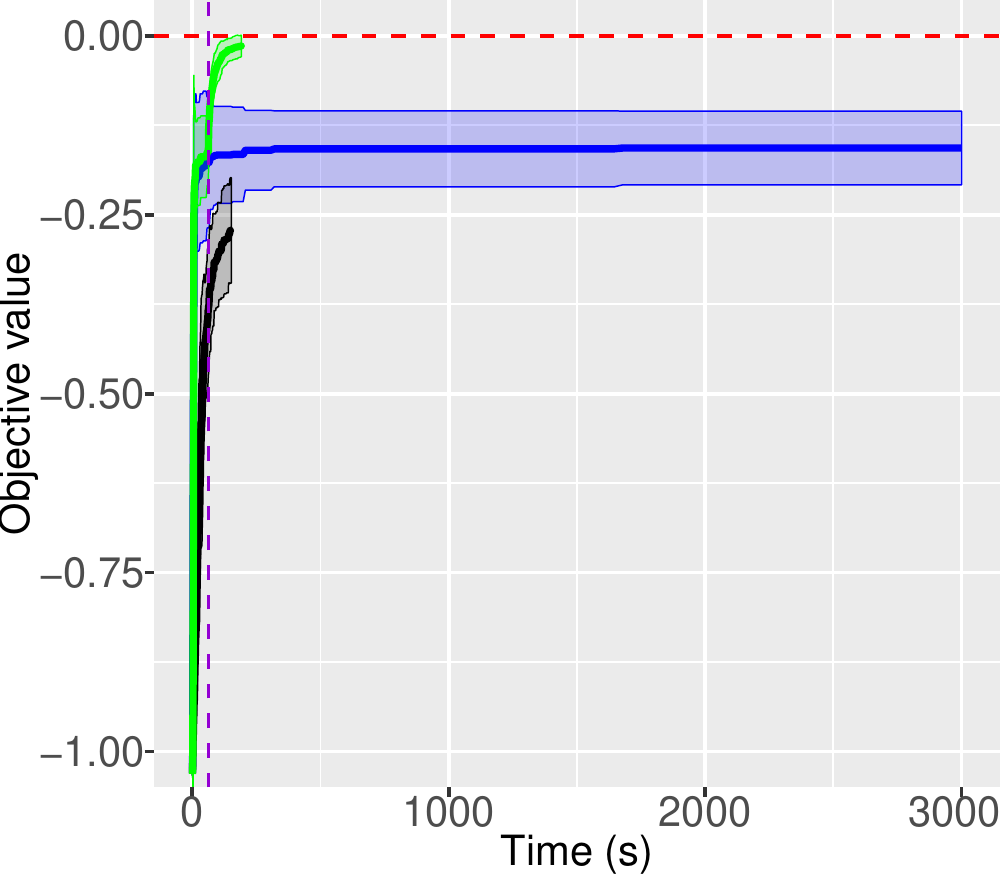}&
        \includegraphics[width=0.49\textwidth]{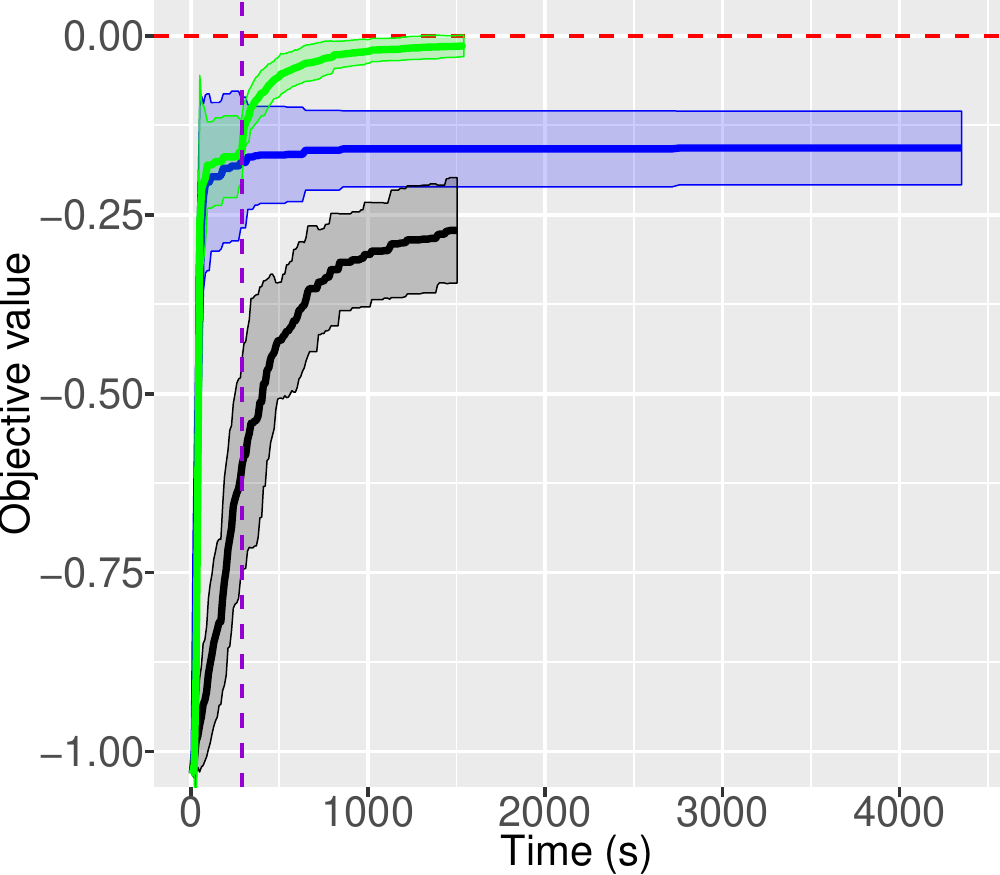}&
        \includegraphics[width=0.49\textwidth]{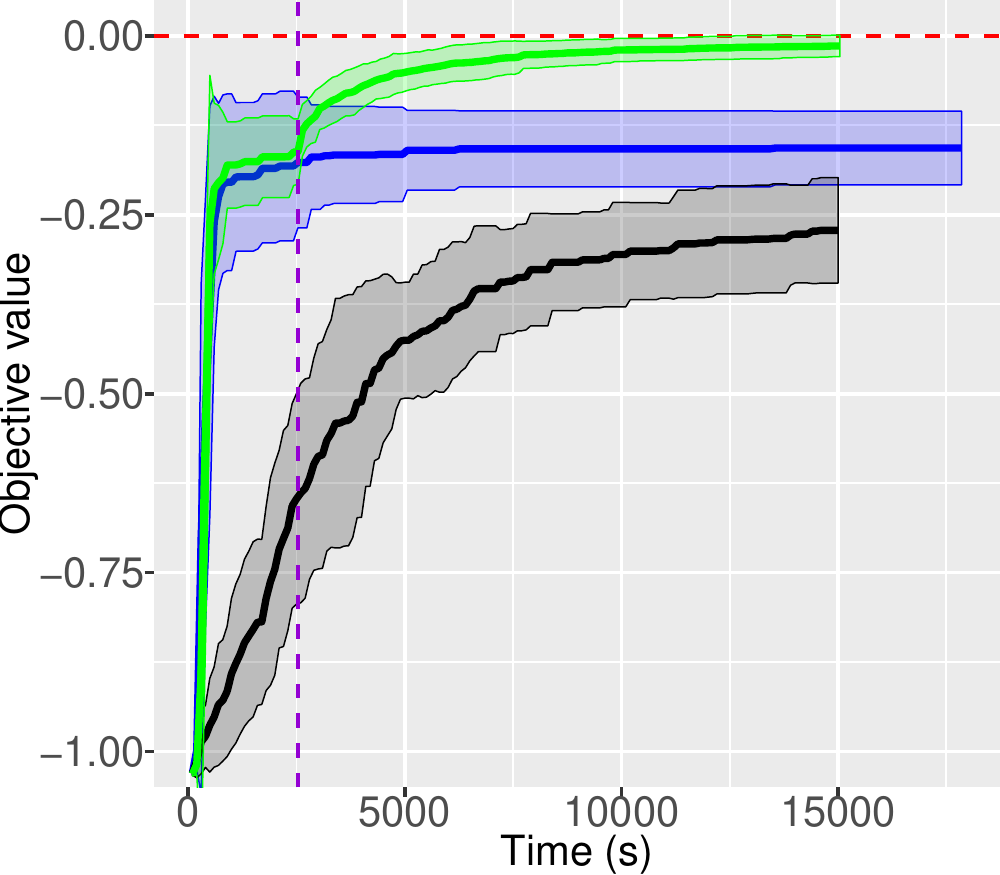}\\
            &   Griewank function    &   \\ \\
        \includegraphics[width=0.49\textwidth]{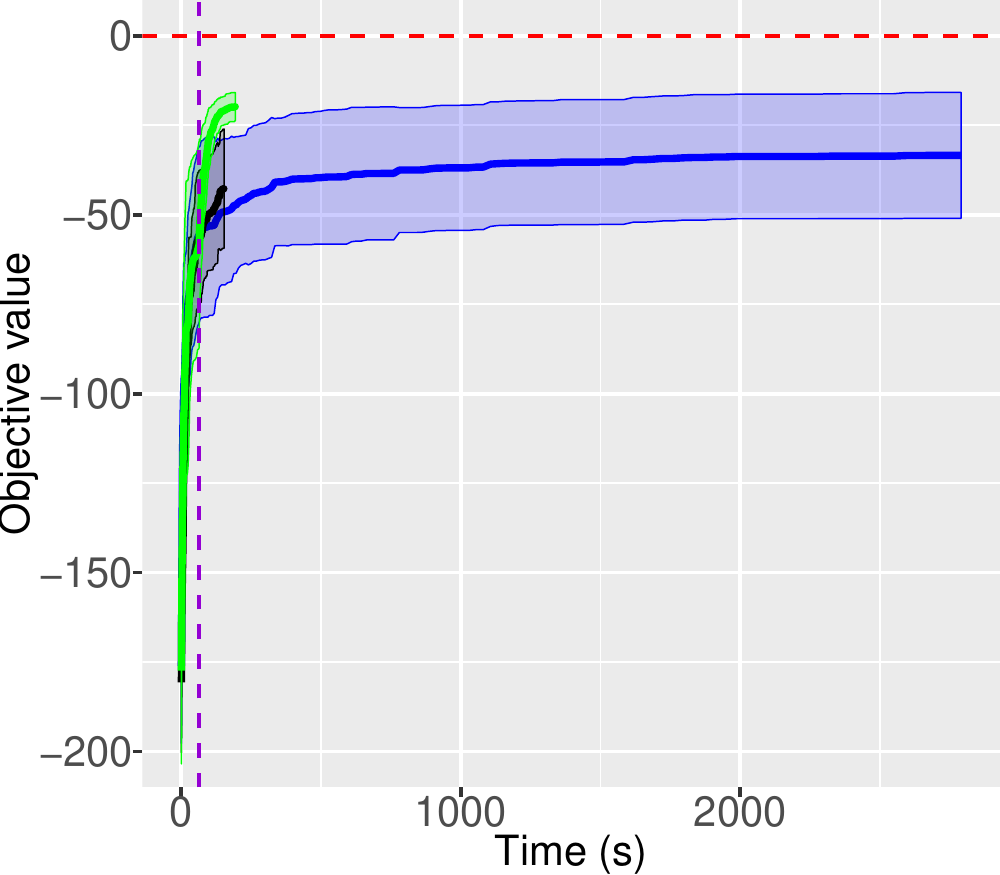}&
        \includegraphics[width=0.49\textwidth]{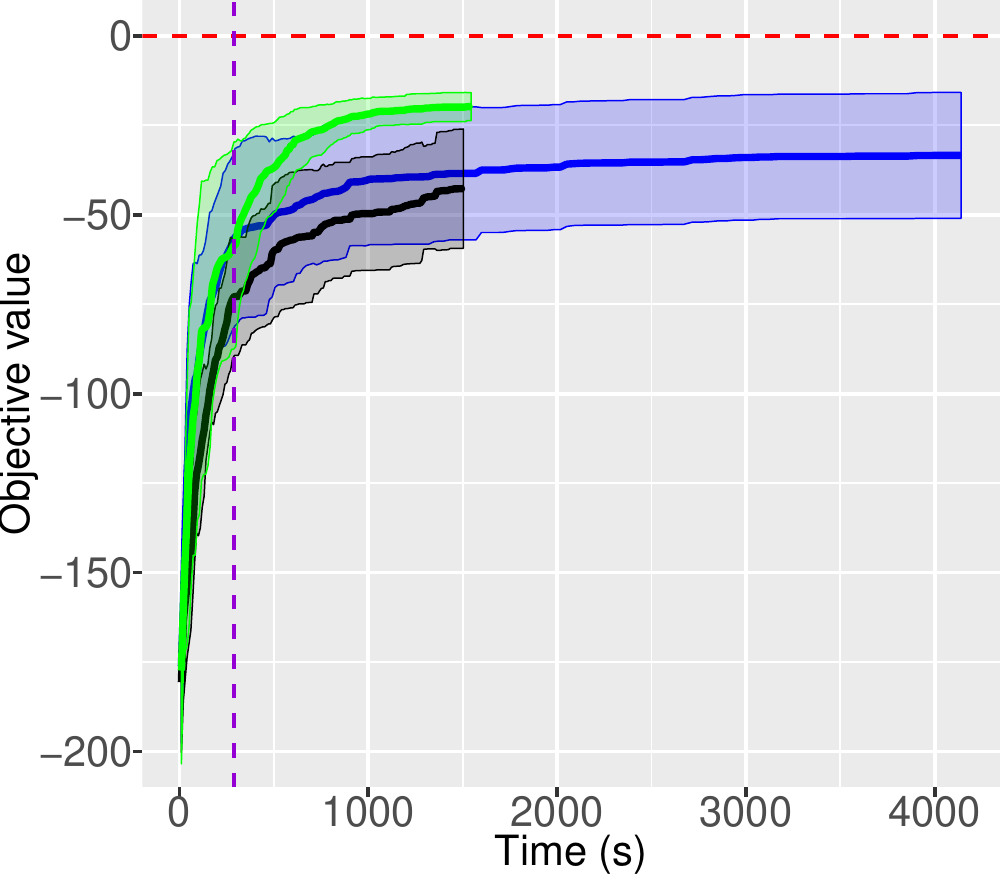}&
        \includegraphics[width=0.49\textwidth]{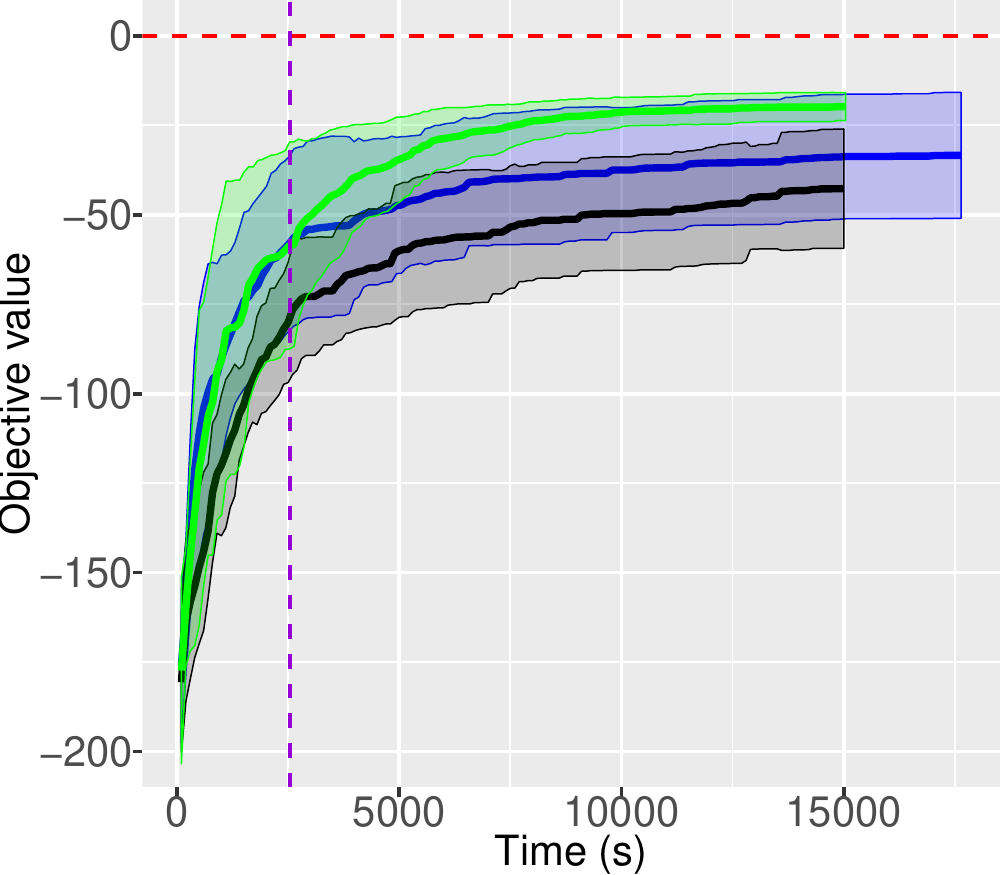}\\
            &   Rastrigin function  &   \\ \\
        \includegraphics[width=0.49\textwidth]{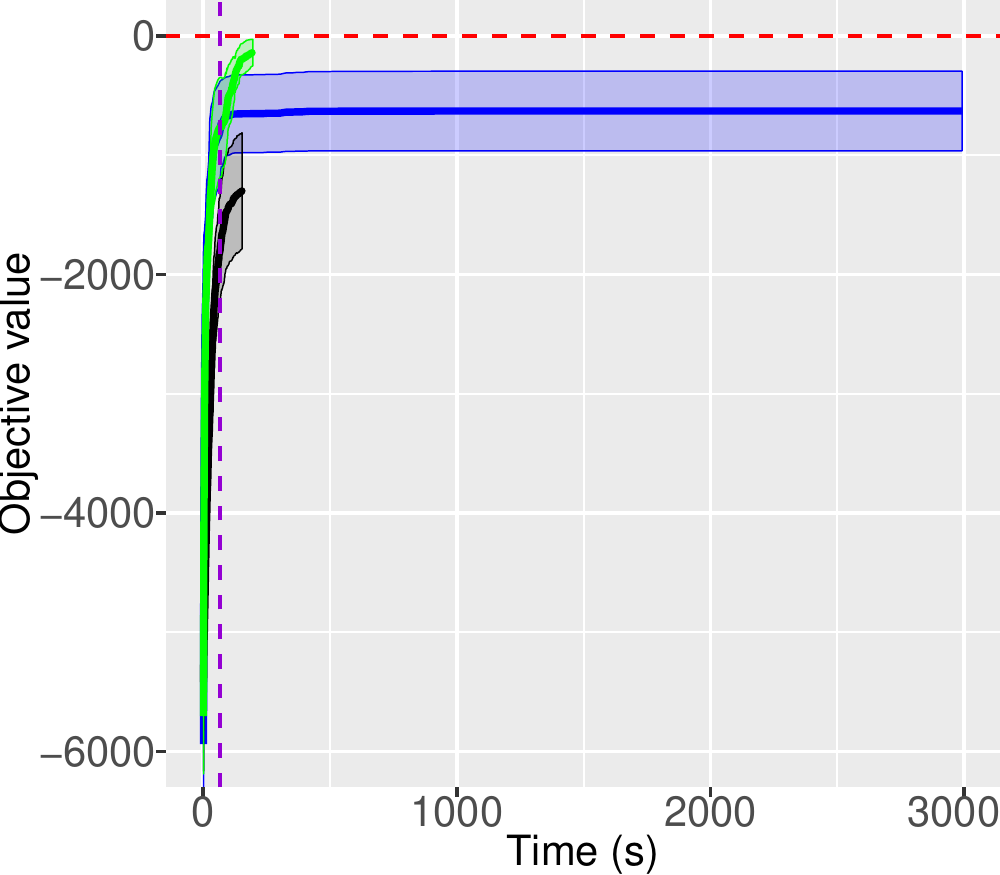}&
        \includegraphics[width=0.49\textwidth]{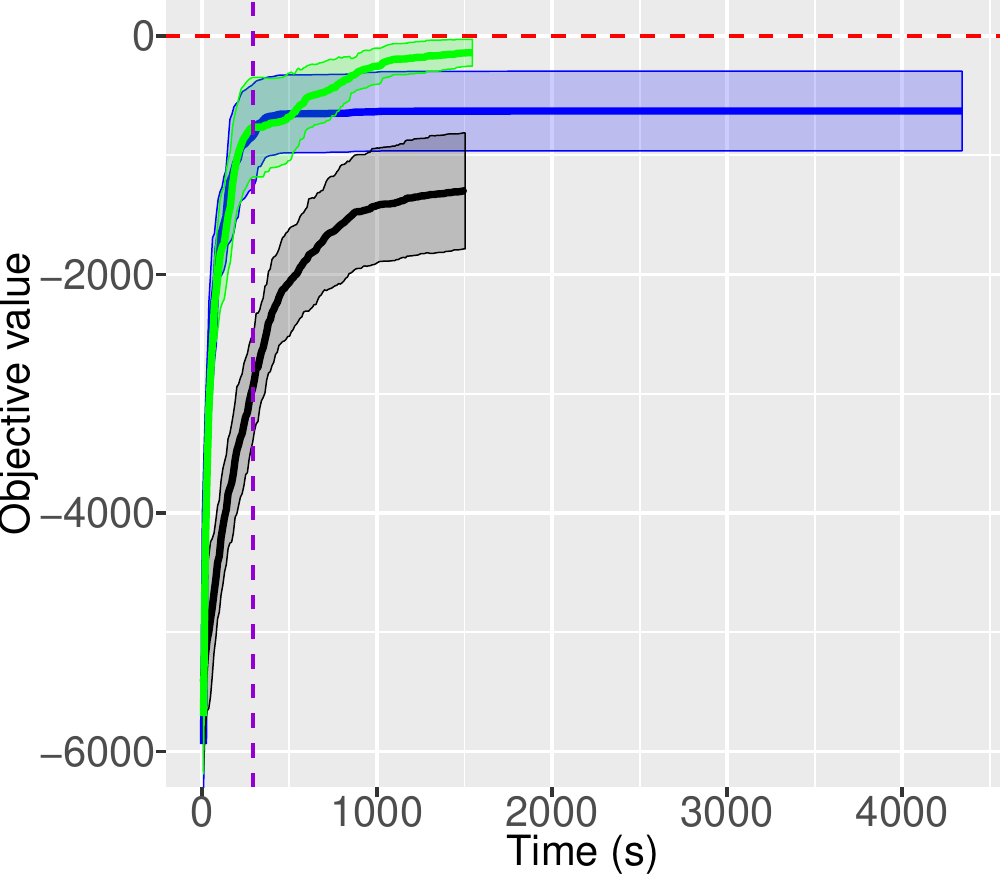}&
        \includegraphics[width=0.49\textwidth]{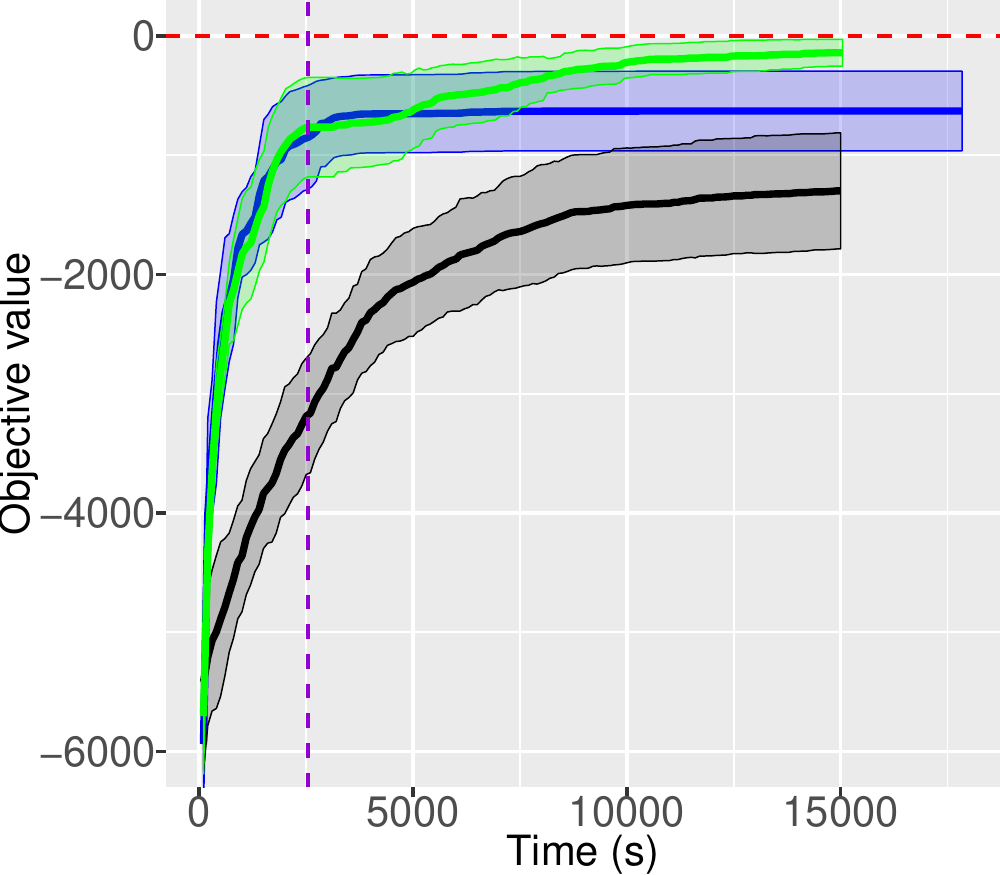}\\
            &   Schwefel function  &   \\
        & \includegraphics[width=0.25\textwidth]{images/ecml_legend} & 
    \end{tabular}
    \end{adjustbox}
    \caption{The mean of best objective values as a function of computation time (in seconds) for Bayesian optimization (blue), evolutionary algorithm (black), and the BEA (green) on the three test functions (top: Griewank, middle: Rastrigin, bottom: Schwefel) with 20 dimensions input. The purple dashed (vertical) lines represent switch points. The red dashed (horizontal) lines are the optimal values of the three test functions. The shadows are the 95.45\% confidence area (two standard deviations).}
    \label{fig:fitness-bo-ea}
\end{figure*}

Finally, 
we run the algorithms on the three objective functions for 1500 iterations and 20 runs per algorithm. 
We observe the algorithm behavior for three different values of evaluation time, $t^e \in \{0.1s, 1s, 10s\}$.
The objective value of the current best solution as a function of total computation time is shown in \autoref{fig:fitness-bo-ea}. 
\par We observe that BO performs better than the EA, and BEA dominates both the EA and BO in terms of objective value.
The computation time of BEA is similar to EA and less than BO, in particular the samll evaluation time, $t^e = 0.1s$.
The computational efficiency, i.e., the gradient of the curves, of BEA is less apparent compared to BO for the case of Schwefel functions with $t^e=10$ just a short time after the switch point. This can be explained by the fact that the switch point was determined heuristically, and possibly it came a bit too early. Nevertheless, both the gains and the objective values of BEA quickly become larger than that of BO as time progresses. Furthermore, we note that the gradient of BEA at the switch point may be less steep than that of EA at the same time. This can be explained by the fact that BEA is closer to the optimal objective value (indicated by the horizontal red line), and thus the objective value is harder to be further improved relative to EA. 

\par In summary, we conclude that BEA outperforms the EA and BO on three benchmark objective functions for the situations of three different evaluation times. Thus, the general idea of selecting which algorithm to run on the basis of expected gain per second (i.e., time efficiency) can indeed be beneficial. 

\section{Application in Robot Learning}
\label{sec:applications}


\par To validate BEA in practice, we apply it in our domain of interest, evolutionary robotics. 
The problem to be solved occurs in a system where robot morphologies evolve and each newborn robot has a new body layout different from its parent(s) \cite{jelisavcic2017real}. 
Then a new robot needs to learn a controller that matches this body and can solve the task(s) in the given application. Thus, the combination of a robot and a task forms a problem instance, where a solution is a controller with high task performance. From the robots' perspective, this is a learning problem. In general, this is a search problem, solvable by any algorithm that can search the space of possible controllers until a solution is found with a high level of task performance. In the sequel, we test BO, the EA, and BEA as learning algorithms for nine test cases of robot learning. 
%
\label{subsec:app_setup}
%
\subsection{Robots and Tasks}
\label{subsubsec:robots}
%
\par
\begin{figure}[!ht]
	\centering
    \begin{adjustbox}{max width=0.9\textwidth}
	\begin{tabular}{c c c}
		\includegraphics[width=0.4\textwidth]{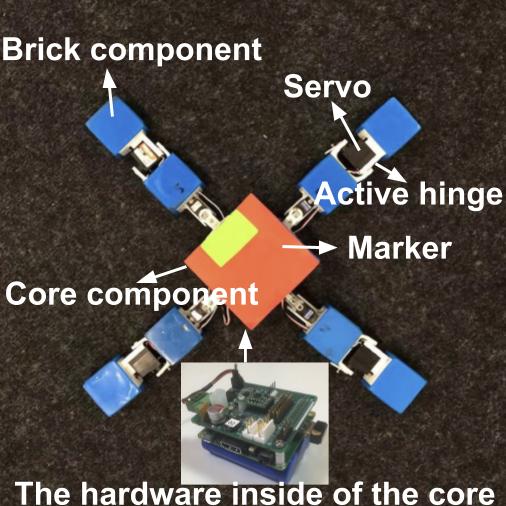}&
		\includegraphics[width=0.4\textwidth]{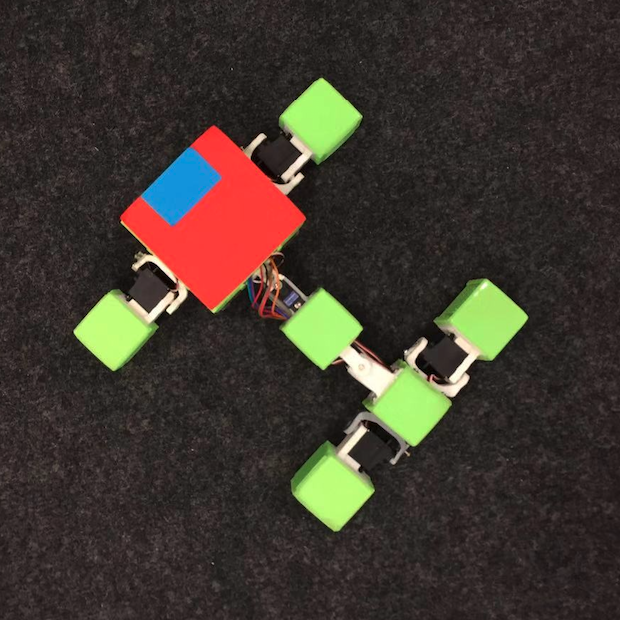}&
		\includegraphics[width=0.4\textwidth]{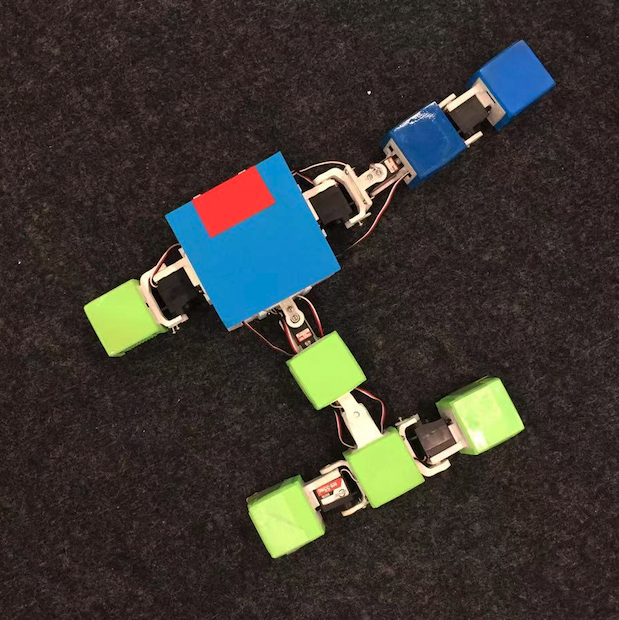} \\
        spider9 & gecko7 & babyA \\
	\end{tabular}
    \end{adjustbox}
	\caption{The prototypes of the three physical modular robots. 
	}
	\label{fig:physical_robots}
\end{figure}

 As a test suite, we chose three modular robots from our previous work, the ``spider9'', the ``gecko7'', and ``babyA'' \cite{lan2018directed,lan2019learning}. Their controllers are CPG based neural networks with a topology determined by the morphologies of the robots. The number of parameters in the controllers that need to be learned depends on the robot shapes. In our case, it is 18, 13, and 16, respectively.
\begin{figure}[!ht]
	\centering
    \begin{adjustbox}{max width=\textwidth}
	\begin{tabular}{c c c}
		\includegraphics[width=0.4\textwidth]{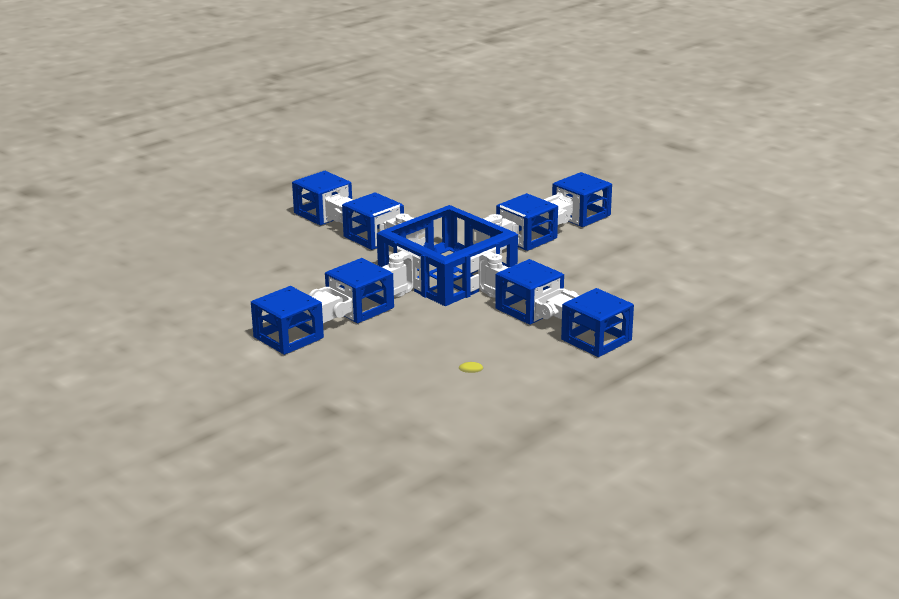}&
		\includegraphics[width=0.4\textwidth]{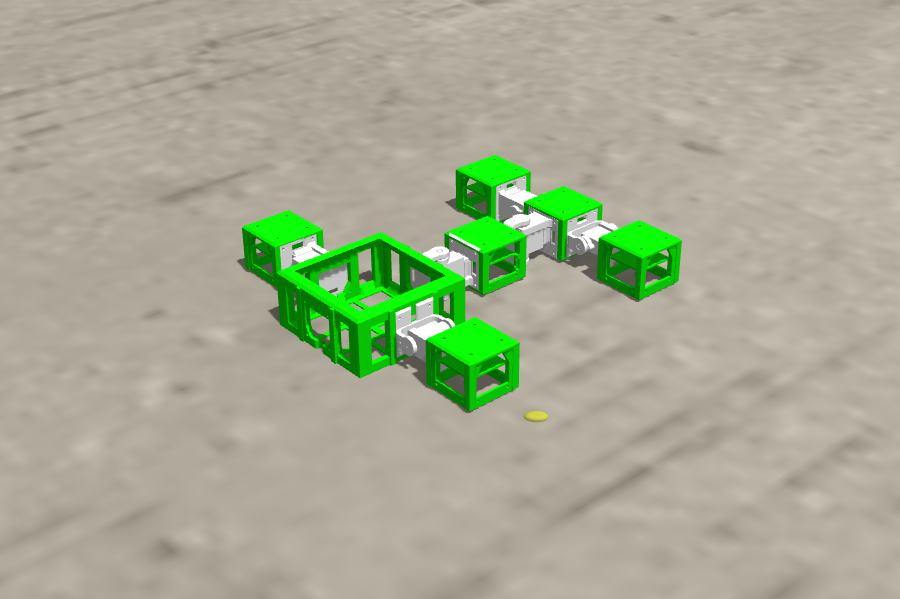}&
		\includegraphics[width=0.4\textwidth]{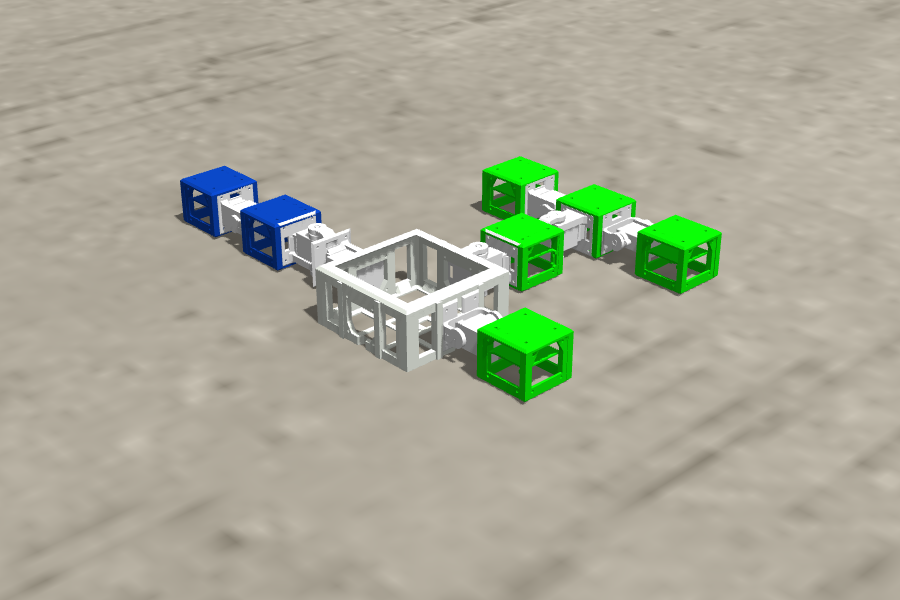} \\
        spider9 & gecko7 & babyA\\
	\end{tabular}
    \end{adjustbox}
	\caption{Illustration of the three simulated modular robots.}
	\label{fig:Morphologies}
\end{figure}
Importantly, we have two incarnations of each robot, a virtual one in our simulation, rf.  \autoref{fig:Morphologies}, and a real one as shown in \autoref{fig:physical_robots}. This allows us to validate the simulated results in the real world. 
%

%

In this paper, we use three tasks: gait learning, directed locomotion, and rotating.
%
%
\textit{Gait learning} is a popular task within evolutionary robotics, where a robot learns to walk as fast as possible without any specific goal \cite{alberto2017current}. 
The task of \textit{directed locomotion} aims to learn controllers for moving as fast as possible towards a given direction. This has more practical value than `just going anywhere' in the task of gait learning, but is hardly investigated in the literature. 
The details of the fitness function we use here can be found in \cite{lan2018directed}.
The third task is \emph{rotating}, where the robot needs to rotate around without changing its location. 
This behaviour is important for seeking points of interest or objects of interest, like a charging station or a possible target, by scanning a neighborhood.
The fitness in this task rewards larger rotation angle and lower displacement over a given test period for the goal of rotating around in the original location. 

%
\subsection{Results with Simulated Robots}
\label{subsec:results_simulated}
%

\par
\begin{figure*}[!ht]
    \centering
    \large
    \begin{adjustbox}{max width=0.91\columnwidth}
    \begin{tabular}{c c c}
        Spider9    &   Gecko7    &  BabyA  \\ 
        \includegraphics[width=0.49\textwidth]{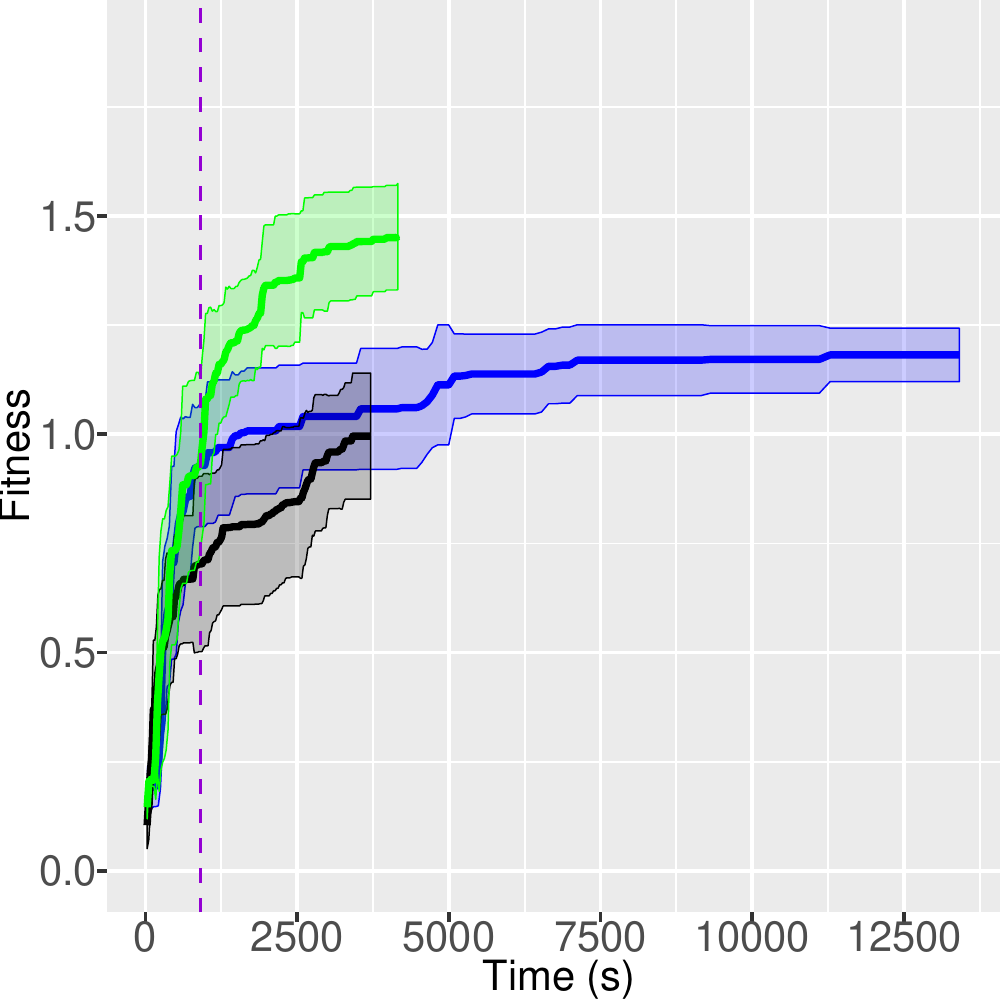}&
        \includegraphics[width=0.49\textwidth]{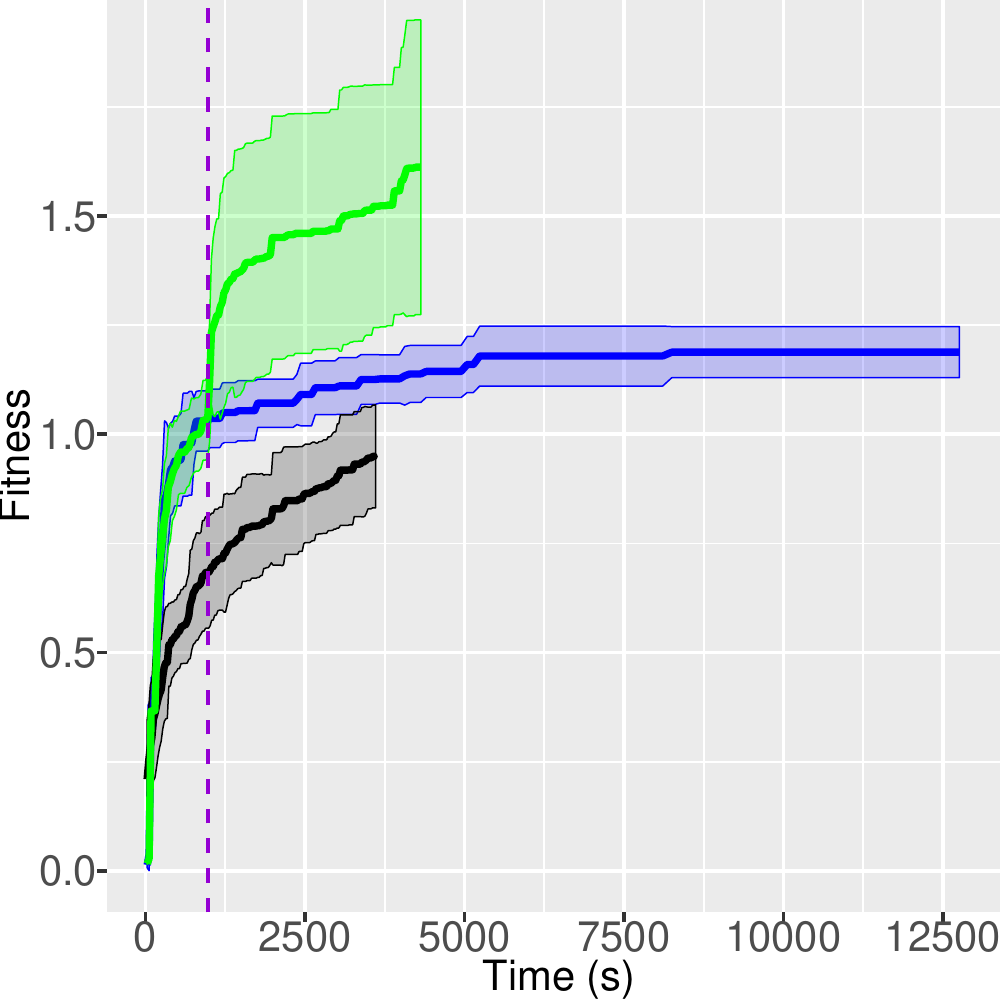}&
        \includegraphics[width=0.49\textwidth]{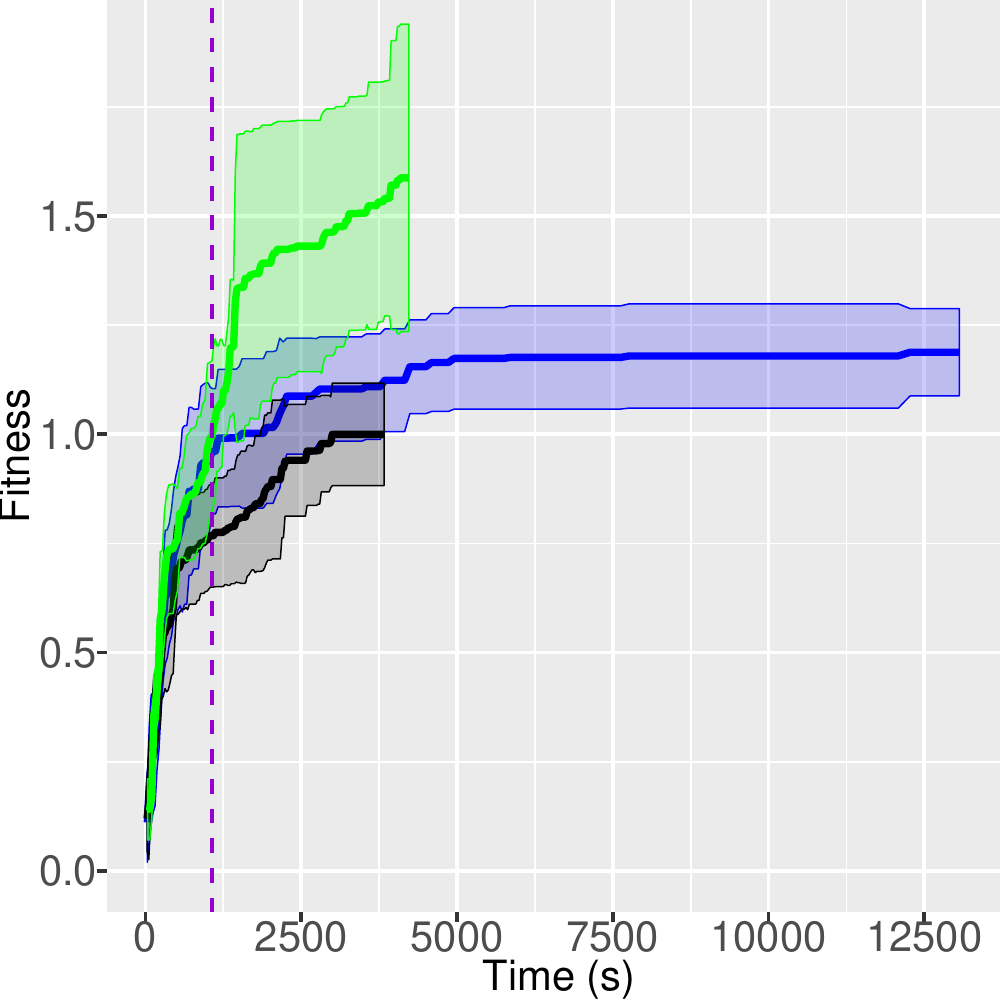}\\
            & Directed locomotion& \\ \\
        \includegraphics[width=0.49\textwidth]{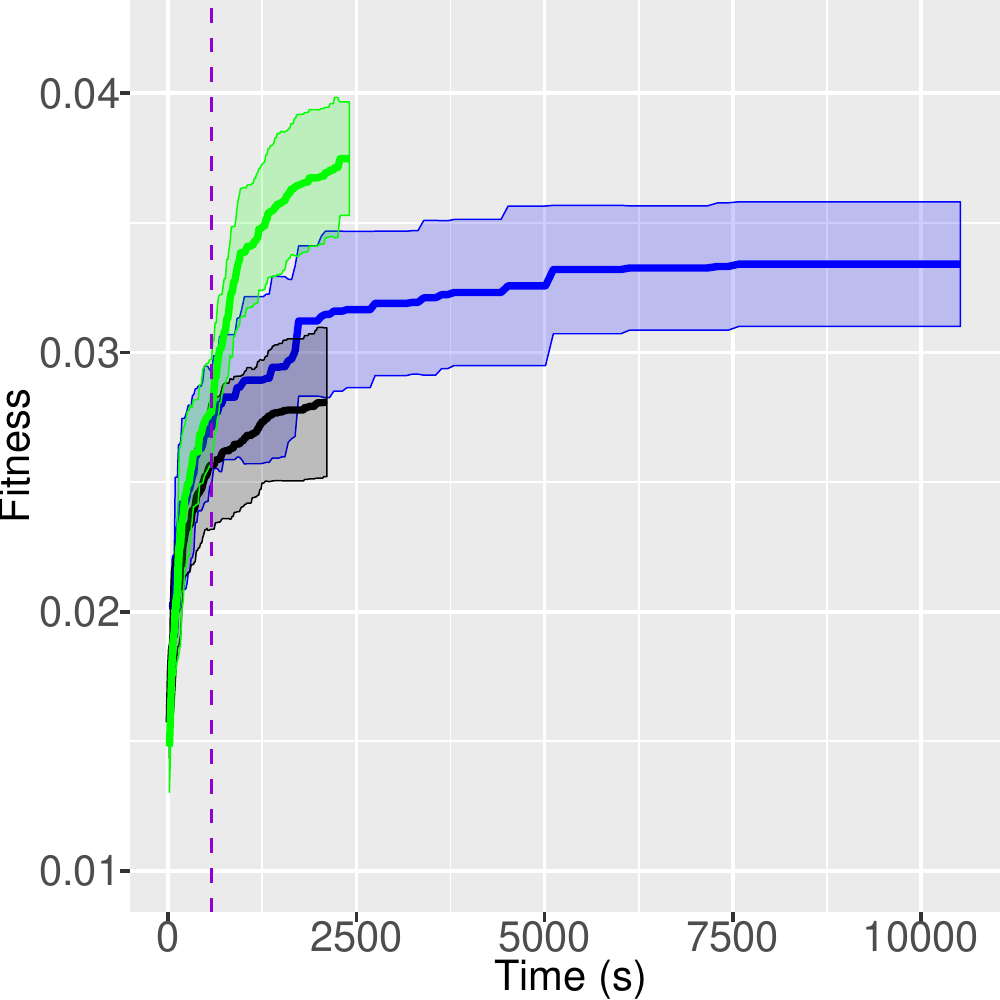}&
        \includegraphics[width=0.49\textwidth]{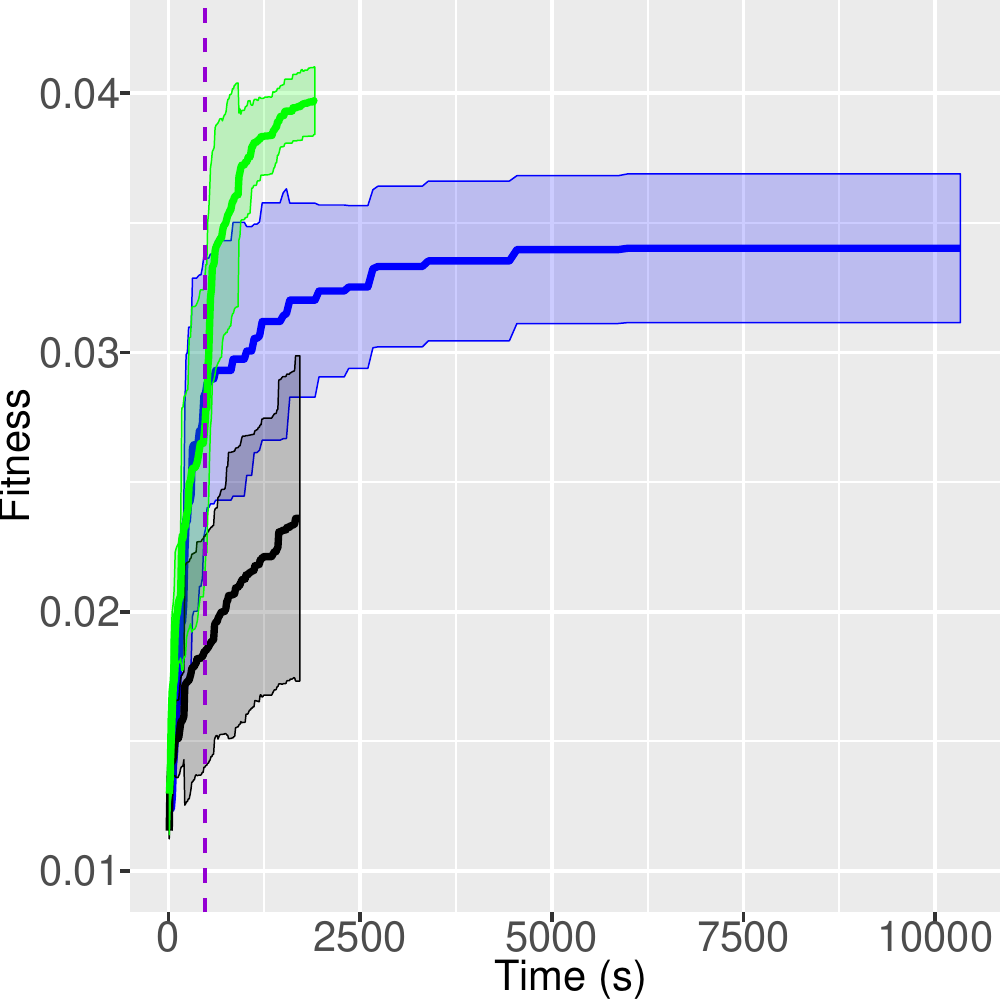}&
        \includegraphics[width=0.49\textwidth]{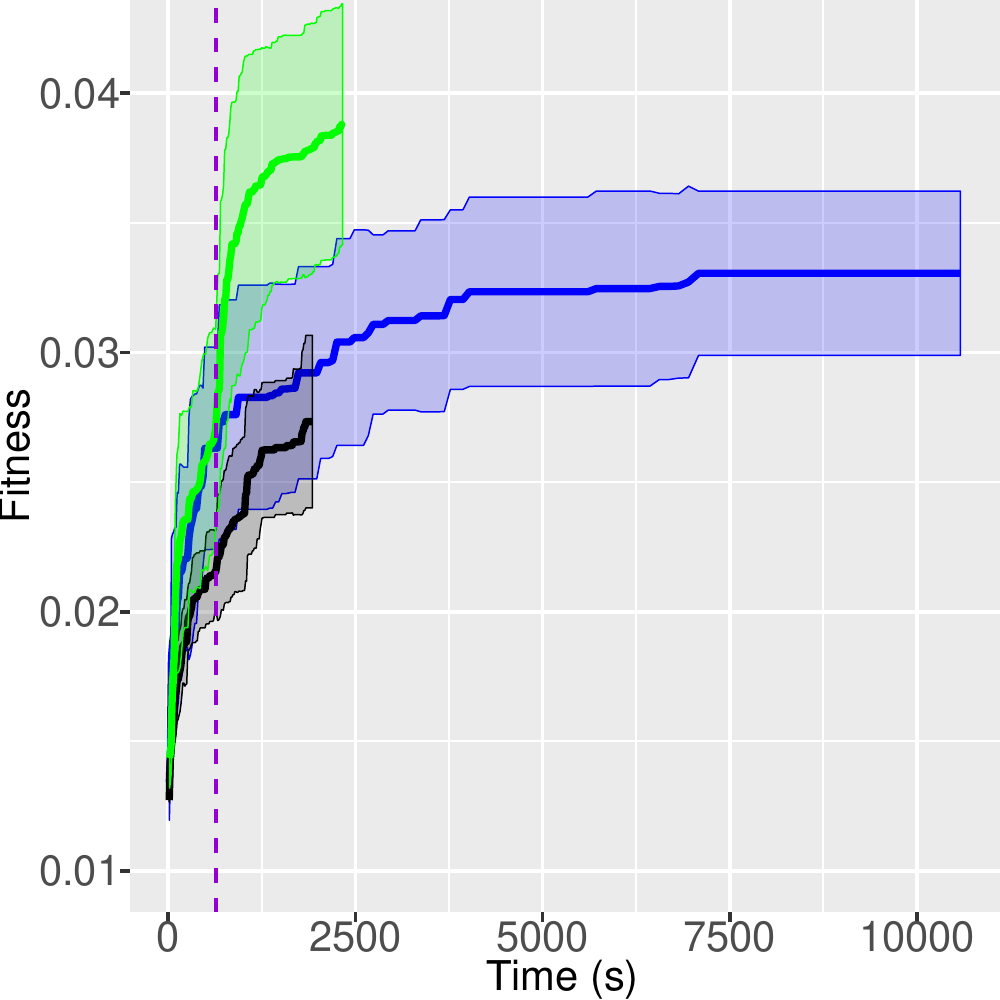}\\
            & Gait learning & \\ \\
        \includegraphics[width=0.49\textwidth]{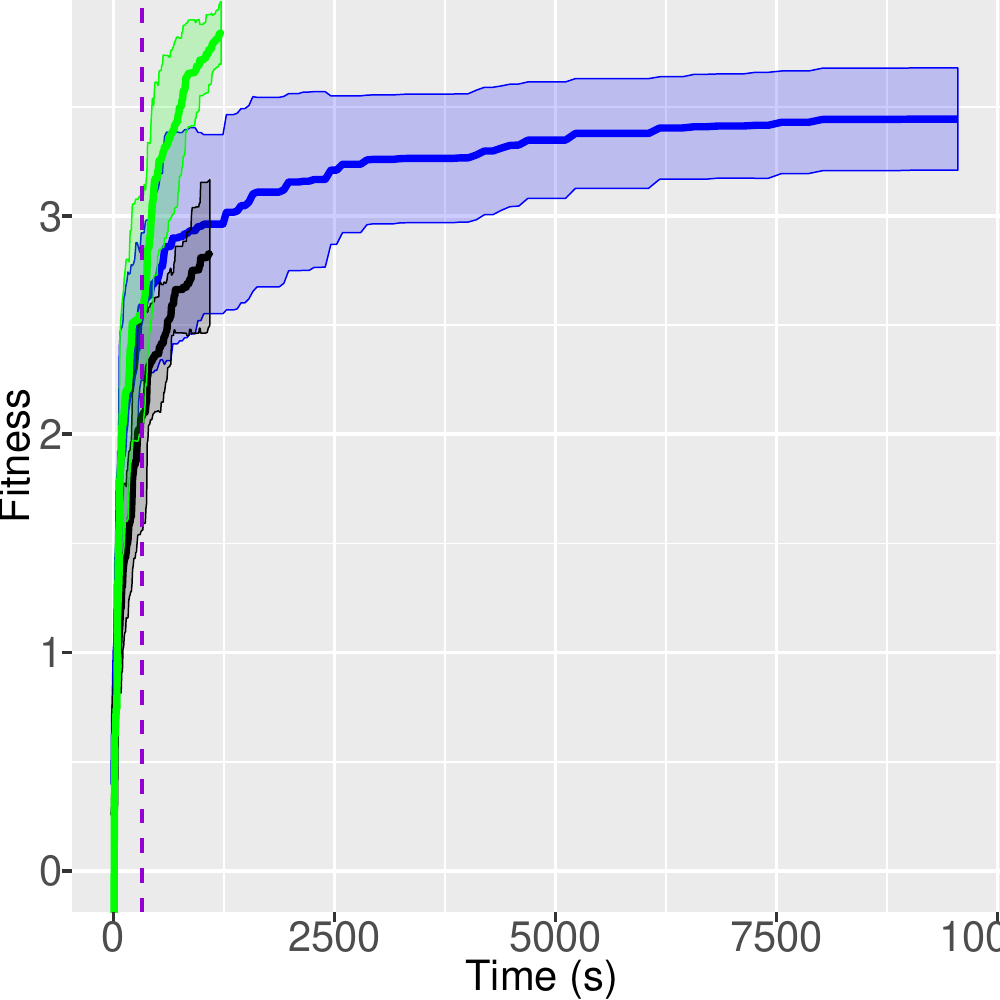}&
        \includegraphics[width=0.49\textwidth]{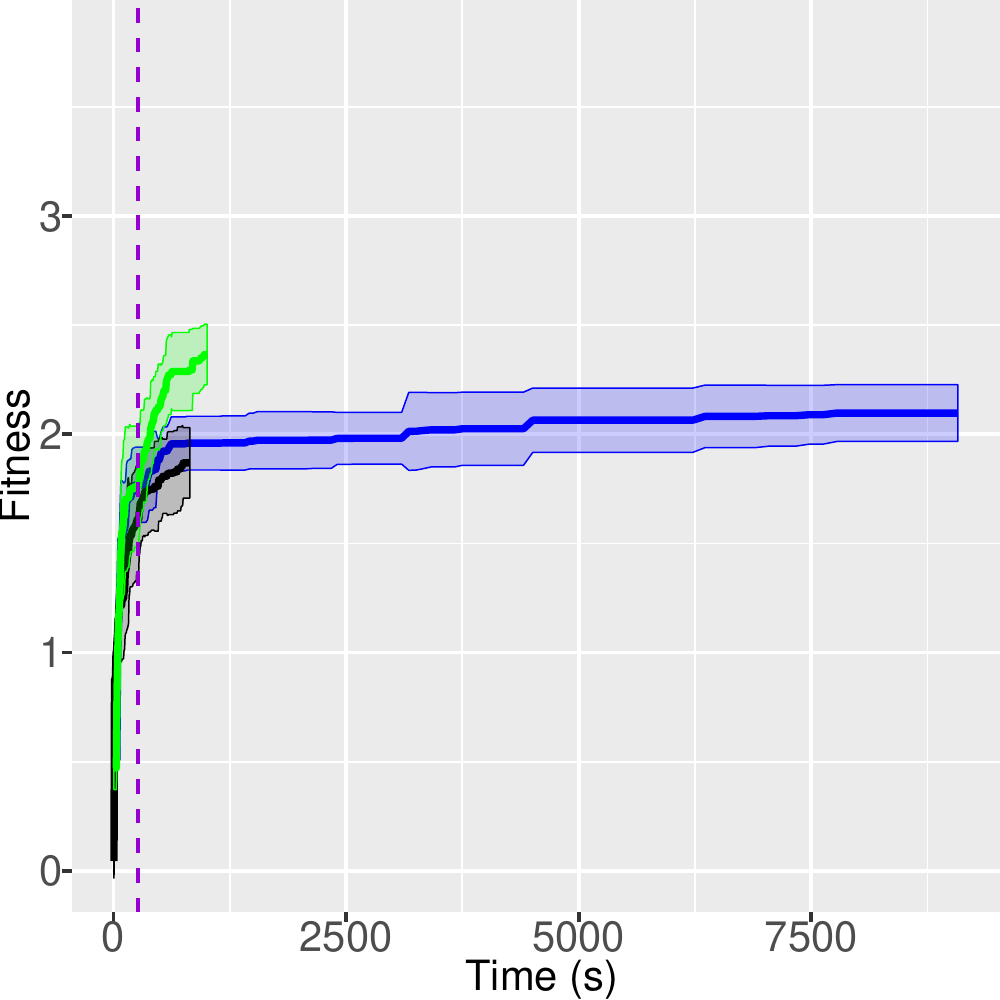}&
        \includegraphics[width=0.49\textwidth]{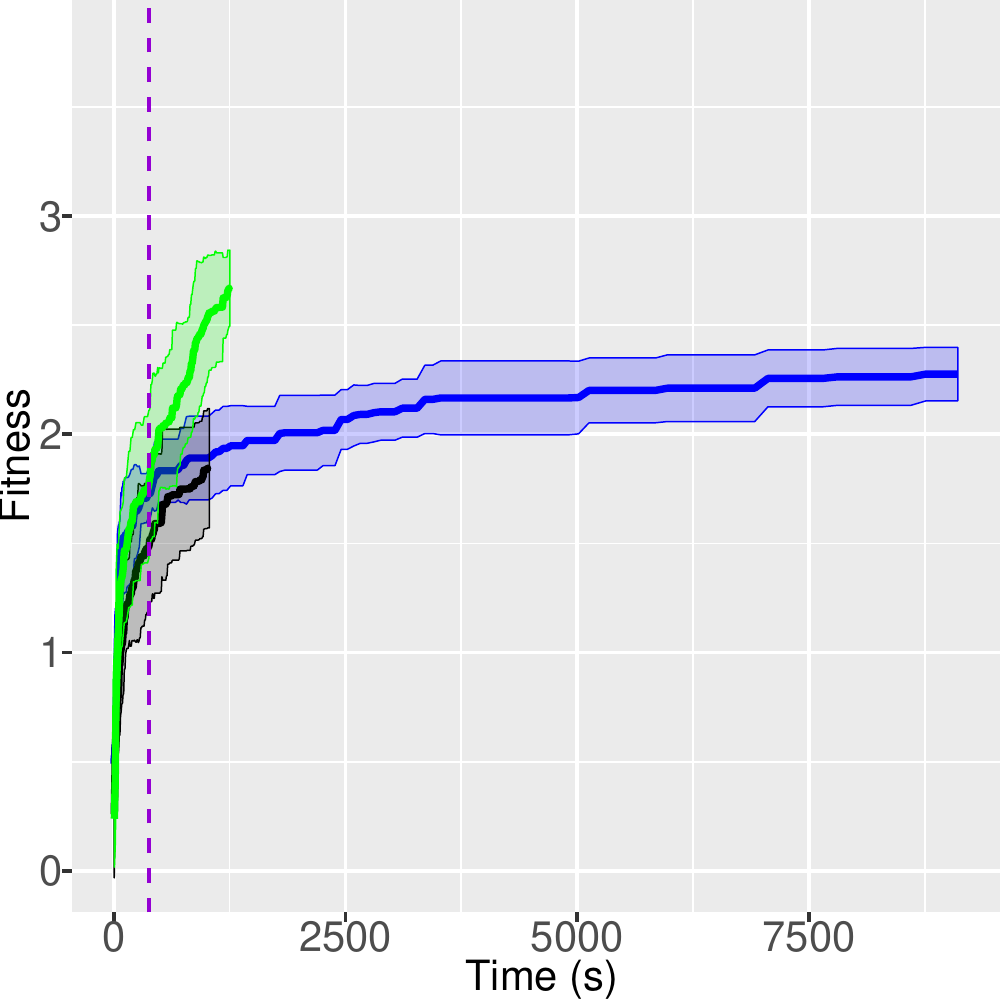}\\
            & Rotating & \\
        & \includegraphics[width=0.25\textwidth]{images/ecml_legend} & 
    \end{tabular}
    \end{adjustbox}
    \caption{Mean of fitness value learned by Bayesian optimization (blue), evolutionary algorithm (black), and the BEA (green) on the three fundamental evolutionary robotics tasks (directed locomotion, gait learning, and rotating) in simulation. The figures in three columns show the results of three tasks for the robots, spider9, gecko7, babyA respectively. The purple dashed lines present the time in switch points. The shadows are the 95.45\% confidence area (two standard deviations).}
    \label{fig:applications}
\end{figure*}

 We apply each algorithm to each of the nine test cases for 10 repetitions in our robot evolution framework \emph{Revolve} \footnote{\url{https://github.com/ci-group/revolve}} \cite{hupkes2018revolve}.
The average best fitness values over the measured computation times are shown in \autoref{fig:applications}.
These curves indicate clear differences among the three algorithms. 
Bayesian optimization (blue) obtains higher fitness by the end of a learning period, and learns much faster than the evolutionary algorithm (black) in the first hundreds of seconds. However, \bo requires much more computation time than the \ea.
Using BEA (green) that switches from BO to the EA after 300 iterations is superior, similarly to the previous experiments on the benchmark test functions. The difference between BEA and BO is significant in all cases.

\par The advantage of BEA with respect to BO and the EA in terms of fitness and computation time is shown in \autoref{tab:comparison}.
For fitness, BEA overtakes BO by $11.5\%$ to $35.6\%$ and the EA by $26.6\%$ to $69.7\%$ over the full run. Regarding computation times, BEA is much faster than BO consuming only one tenth to one third of the time to finish and a bit slower than the EA needing about $10\%$ to $20\%$ more time. 

\begin{figure}[!htbp]
	\centering
    \begin{adjustbox}{max width=0.8\textwidth}
	\begin{tabular}{c}
		\includegraphics[width=0.95\textwidth]{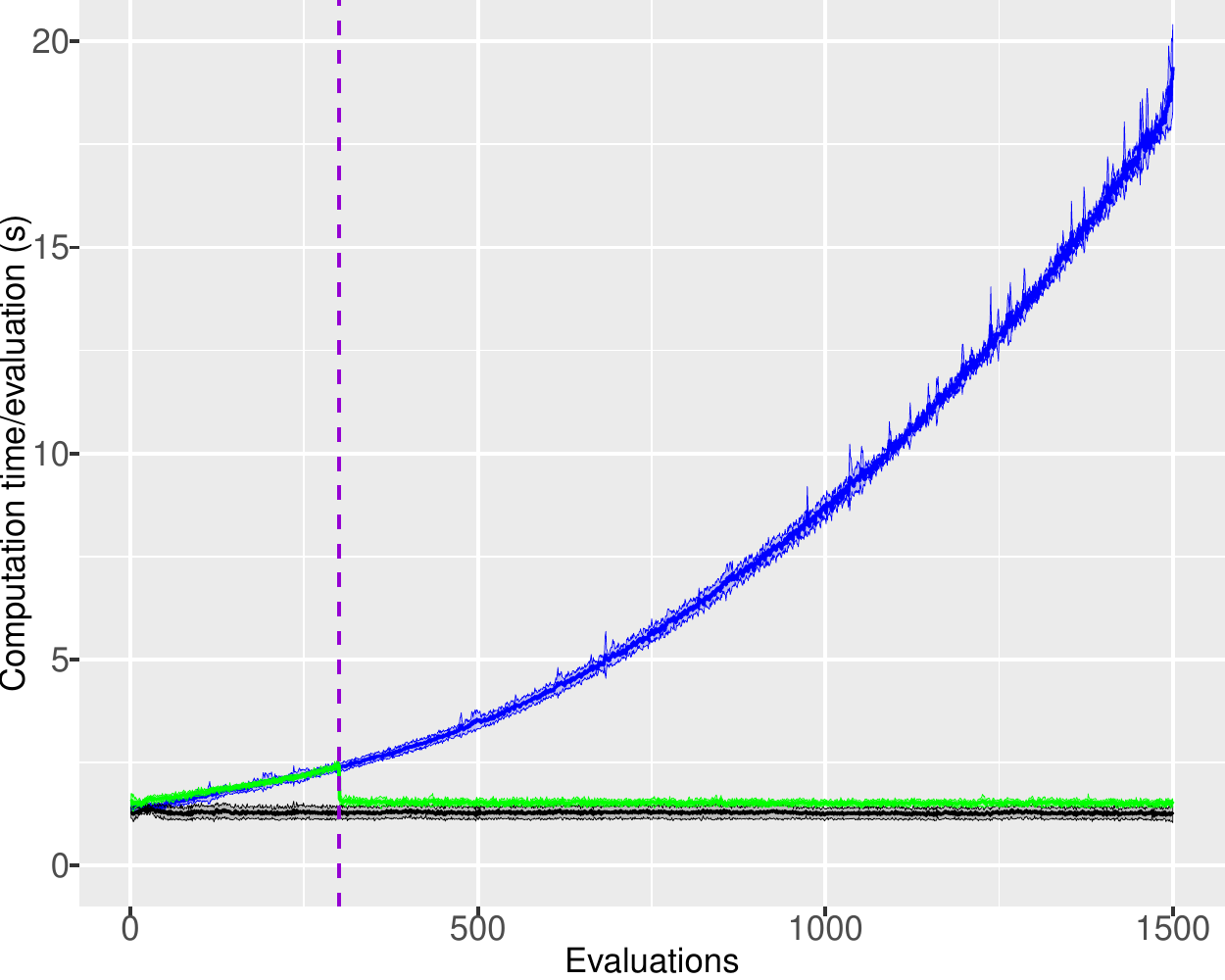} \\
		\includegraphics[width=0.35\textwidth]{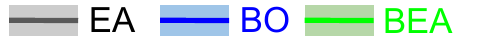}
	\end{tabular}
    \end{adjustbox}
	\caption{The mean and standard deviation of computation time for BO, the EA, and BEA over 10 repetitions for the task of gait learning on the robot ``babyA'', shown by the blue, black, and green lines, respectively. The purple line indicates the switch point in BEA after 300 evaluations. 
	}
	\label{fig:ctime}
\end{figure}

%
\begin{table*}[!hbp]
    \centering
    \small
    \setlength\tabcolsep{6pt} 
    \begin{tabular}{l|l|ccc|ccc|ccc}
    \toprule
    \multicolumn{2}{c|}{tasks$\rightarrow$} & \multicolumn{3}{c|}{Directed locomotion} 
                  & \multicolumn{3}{c|}{Gait learning} 
                  & \multicolumn{3}{c}{Rotating} \\
    \multicolumn{2}{c|}{robots$\rightarrow$}  &  \multicolumn{1}{c}{spider9} & \multicolumn{1}{c}{gecko7} & \multicolumn{1}{c}{babyA} & \multicolumn{1}{|c}{spider9} & \multicolumn{1}{c}{gecko7} & \multicolumn{1}{c}{babyA} & \multicolumn{1}{|c}{spider9} & \multicolumn{1}{c}{gecko7} & \multicolumn{1}{c}{babyA} \\ \midrule
    best fitness & BO   & 122.9\% & \textbf{135.6\%} & 133.6\% & 112.2\% & 116.7\%& 117.4\% & \textbf{111.5\%} & 112.8\% & 117.3\%\\
    of BEA w.r.t. & EA  & 145.9\% & \textbf{169.7\%} & 158.7\% & 133.5\% & 168.3\% & 141.9\% & 135.4\% & \textbf{126.6\%} & 144.7\% \\ \cline{1-11}
    comp. time  & BO   & 31.0\% &  \textbf{33.8\%} & 32.4\% & 22.8\% & 18.5\% & 22.0\% & 12.8\% & \textbf{11.1\%} & 13.8\%\\
    of BEA w.r.t. & EA  & 112.2\% & 119.6\% & \textbf{110.4\%} & 114.0\% & 111.3\% & 121.0\% & 111.7\% & \textbf{122.7\%} & 121.2\%\\ \bottomrule
    \end{tabular}
    \caption{Simulation based comparison. Upper half: The performance of BEA in terms of fitness over a full run w.r.t. BO and EA defined by BEA/BO and BEA/EA respectively. Lower half: BEA vs. BO and the EA in terms of computation time defined as BEA/BO and BEA/EA respectively.}
    \label{tab:comparison}
\end{table*}
%
To provide further insights and to verify the cubic time complexity of Bayesian optimization, we plot the computation time (in seconds) over the consecutive evaluations. Here we do not present all nine test cases, only the one for gait learning on the robot ``babyA'' in \autoref{fig:ctime}, in which the evaluation time is near to 1 seconds wall-clock time; the other ones show almost identical trends.
The blue line shows that computation times of BO increase rapidly. The black line exhibits small fluctuations of more or less constant computation times for the evolutionary algorithm and the green curve for BEA follows the blue curve in the first stage and the black line after the switch point (purple line) as we expected. This clearly demonstrates that BEA cures the time complexity problem of BO, providing the best of both worlds.

\par
\begin{figure}[!ht]
	\centering
    \begin{adjustbox}{max width=0.8\textwidth}
	\begin{tabular}{c}
		\includegraphics[width=0.99\textwidth]{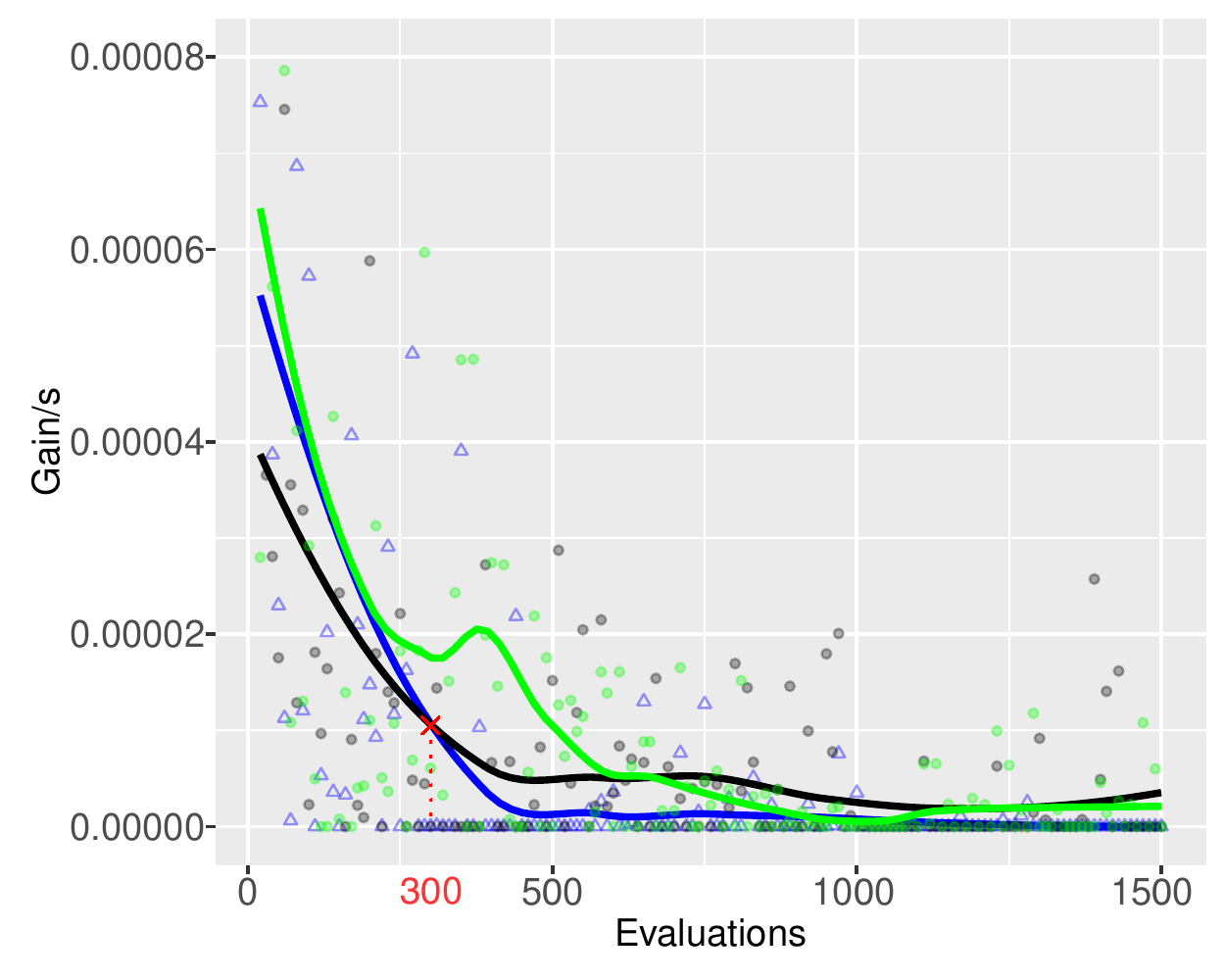} \\	\includegraphics[width=0.35\textwidth]{images/ecml_legend.pdf}
	\end{tabular}
    \end{adjustbox}
	\caption{The gain per second for BO, the EA, and BEA for the task of gait learning on the robot ``babyA''. The solid lines show the smoothed conditional mean (in blue, black, and green, respectively). The red line and the cross indicate the switch point in BEA after 300 evaluations.
	}
	\label{fig:gain_applications}
\end{figure}

 As outlined in \autoref{sec:efficiency}, the gain per second ($\mathcal{G}_{(k,i)}$) can be used to measure the time efficiency of optimization algorithms. To investigate how this works in practice, we present the plots that show how time efficiency changes over consecutive evaluations for the task of gait learning on the robot "babyA" in \autoref{fig:gain_applications}.
As we can see, BO (the blue curve) achieves higher gains per second than the EA (in black) in the beginning.
However, the gain per second for BO decreases rapidly and drops under that of the EA after the switch point. From about 500 evaluations the BO gains are per second are near zero. Compared to BO, the EA shows a more gradual decline. 
As expected, BEA exhibits a behavior very similar to BO in the first stage, although the curves of BEA and BO are not exactly same because of the smoothed conditional mean. After switching to the EA, BEA shows a sharp increase of gain per second in the first hundreds of evaluations. This indicates that the initial population BO seeded the EA with was indeed `useful'. After the first hundreds of evaluations in the third stage, the gains for BEA start dropping and become lower than the gain per second of the EA. This is not surprising, because the fitness values for BEA are closer to the optimum and are harder to improve.
%

\subsection{Results with Real Robots}
\label{subsec:results_real}
%

\par The proof of the pudding is in the eating. To further demonstrate the merits of BEA as a method to learn good robot controllers, we compare the performance of the best controllers obtained by BO and BEA on each of nine test cases in the real world.\footnote{We omit the EA from this comparison, because the quality of its solutions is consistently lower than the quality obtained by BO and BEA.} To this end, we install these controllers on the real robots and run them for a test period in a flat arena with an overhead camera localization system that measures the movements, logs the trajectories, and captures a video.\footnote{The video of the real robot behaviours with the best controllers learned by BEA for nine test cases is shown in \url{https://youtu.be/V8Y8Jhnzx_w}} The duration of the test period differs each task: 60, 30, and 15 seconds for directed locomotion, gait learning, and rotating, respectively. The results of these tests, averaged over three repetitions, are shown in \autoref{tab:comparison_real}. These fitness values, calculated by actual measurements, provide convincing evidence for the advantage of BEA. In eight out of the nine test cases the BEA-powered robot behavior is better --and sometimes far better-- than the behavior induced by the BO-optimized controller. 

%

\begin{table*}[!htbp]
    \centering
    \small
    \setlength\tabcolsep{6pt} 
    \begin{tabular}{l|l|ccc|ccc|ccc}
    \toprule
    \multicolumn{2}{c|}{tasks $\rightarrow$} & \multicolumn{3}{c|}{Directed locomotion} 
                  & \multicolumn{3}{c|}{Gait learning} 
                  & \multicolumn{3}{c}{Rotating} \\
    \multicolumn{2}{c|}{robots $\rightarrow$}  &  \multicolumn{1}{c}{spider9} & \multicolumn{1}{c}{gecko7} & \multicolumn{1}{c}{babyA} & \multicolumn{1}{|c}{spider9} & \multicolumn{1}{c}{gecko7} & \multicolumn{1}{c}{babyA} & \multicolumn{1}{|c}{spider9} & \multicolumn{1}{c}{gecko7} & \multicolumn{1}{c}{babyA} \\ \midrule
    \multirow{2}{*}{ fitness} & \quad BO & 0.82 & 0.66 & 0.83 & 0.032 & 0.031 & 0.035 & 1.99 & 0.90 & 0.52 \\
        & \quad BEA & 0.70 & 1.77 & 1.48 & 0.035 & 0.036 & 0.036 & 2.91 & 1.49 & 0.87 \\ \midrule
    \multicolumn{2}{l|}{ratio of BEA w.r.t. BO} & 85.4\% & 268.5\% & 178.5\% & 109.1\% & 108.2\% & 103.4\% & 146.5\% & 165.8\% & 168.2\%\\ \bottomrule
    \end{tabular}
    \caption{Real world performance of the BO- and BEA-learned controllers. These fitness values are calculated by actual measurements during the test periods. The ratio of BEA w.r.t. BO is defined as BEA/BO.}
    \label{tab:comparison_real}
\end{table*}
%

\par
\begin{figure*}[!ht]
\centering
\Large
\begin{adjustbox}{max width=0.99\textwidth}
	\begin{tabular}{c c c}
		\includegraphics[width=0.5\textwidth]{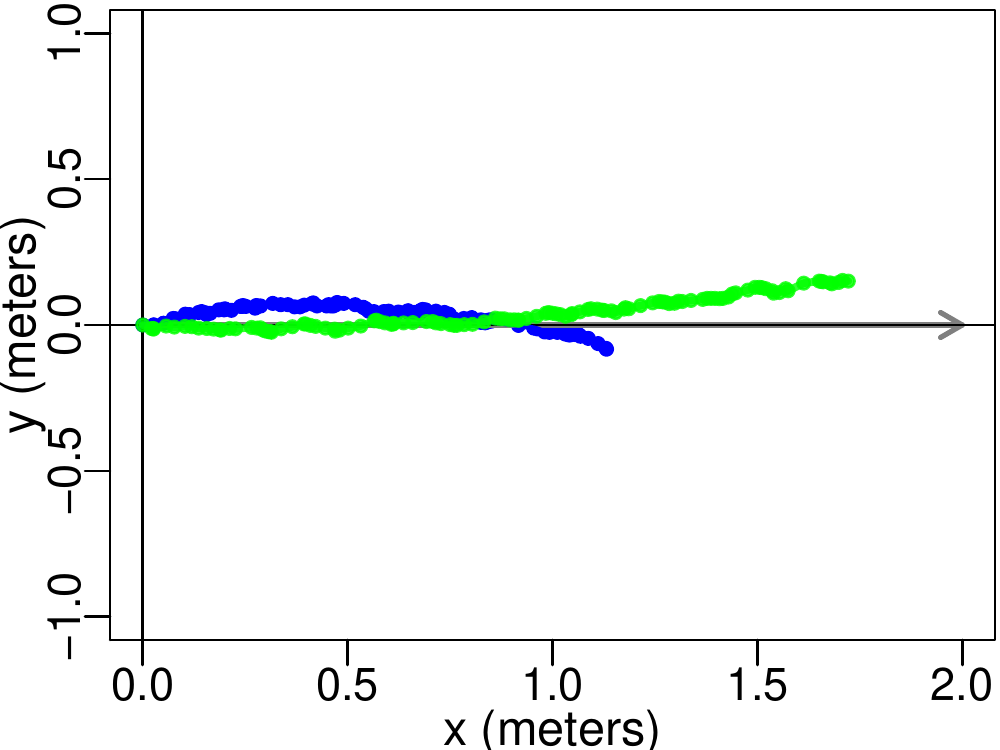} & 
		\includegraphics[width=0.5\textwidth]{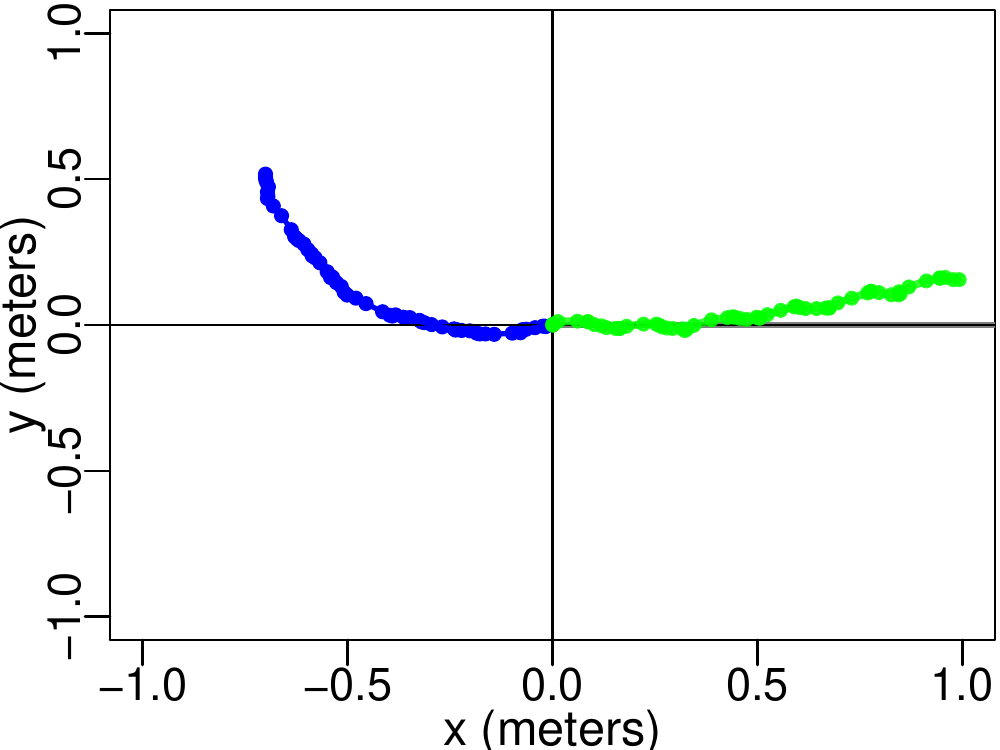} &
		\includegraphics[width=0.5\textwidth]{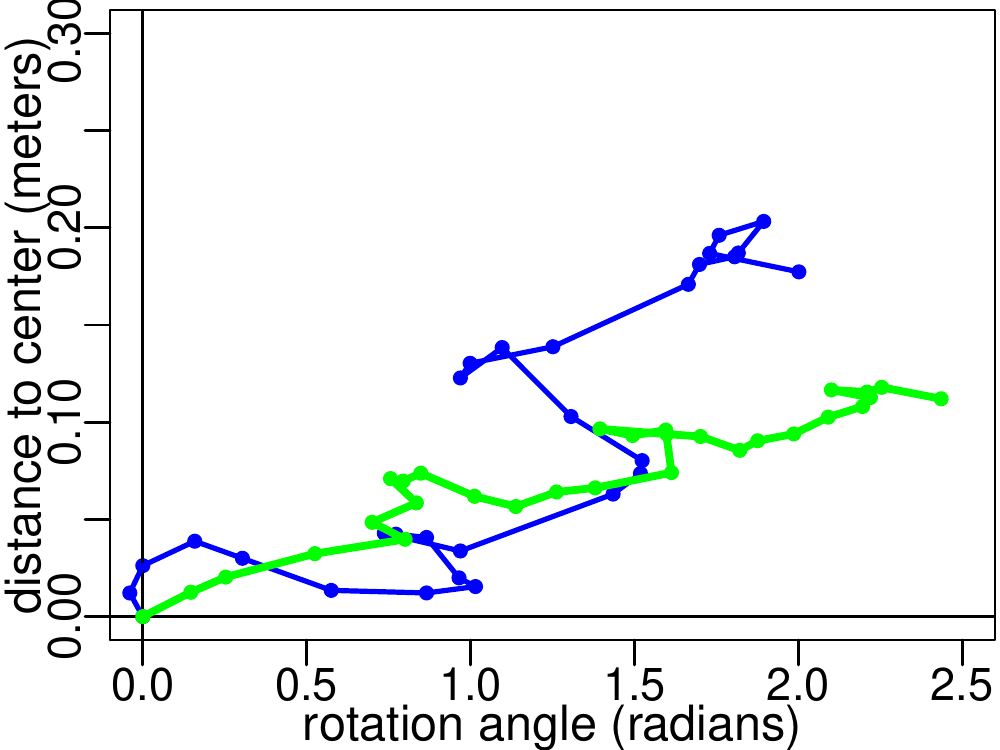} \\
        Directed locomotion (in 60 seconds) & Gait learning (in 30 seconds) & Rotating (in 15 seconds) \\
                & \includegraphics[width=0.17\textwidth]{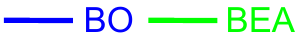} & 

	\end{tabular}
\end{adjustbox}
\caption{\label{fig:paths_real} 
Plots showing the behavior of the real robot ``babyA''. Controllers learned by BO / BEA are shown in blue / green respectively. See the text for detailed explanation.}
\end{figure*}

 The fitness values shown in \autoref{tab:comparison_real} provide a comparison between the algorithms by abstract numbers that are hard to interpret in terms of the actual robot behavior. To demonstrate the latter, we display the average trajectories of ``babyA'' with the best controllers learned by BEA and BO. The plot on the left hand side of \autoref{fig:paths_real} shows the averaged trajectories for directed locomotion. The aim is to move in a straight line from the origin $(0, 0)$ following the target direction indicated by the arrow. As we can see, the robot moves in the right direction with both controllers, but covers a larger distance (1.72m) when it uses the controller delivered by BEA (green line). The plot in the middle belongs to the gait learning task, where the direction is not important, the robot only aims to move as far as possible from the starting point. Also here, the best controller learned by BEA induces better behavior than the one delivered by BO. Finally, the figure on the right hand side shows the average rotation angle and the distance to center for the task of rotating. By the nature of this task, the curve for a good behavior stays low (robot remains close to the center) and gets far to the right (the robot makes a rotation of many radians). The BEA produced controller is better in this case too; ``babyA'' rotates more than 2 radians and moves just 10 centimeters. The best controller learned by BO is rotating less and moves farther away from the starting point. In summary, we conclude that the best controllers learned by BEA and BO conduct the three tasks on real robots successfully, but the controllers learned by BEA are better.
%


\section{Conclusion}
\label{sec:conclusion}
In this paper we addressed an issue about how the performance of generate-and-test search algorithms should be measured. We argued that data efficiency is not sufficient and we should also consider time efficiency. We proposed to calculate the time efficiency of search algorithms as the gain in objective function value per unit of time. We compared the time efficiency of Bayesian optimization and an evolutionary algorithm and observed that BO becomes less time efficient than the EA after a while due to BO's polynomial complexity in the number of iterations (i.e., the number of previously evaluated candidate solutions). 

To deal with data efficiency and time efficiency at the same time, we proposed a new algorithm, BEA, that combines the best of two worlds by switching from BO to an EA during the search process. This algorithm benefits from the data efficiency and relatively low computational cost of BO in the beginning and enjoys the better time efficiency of the EA in later iterations. In order to switch from BO to the EA effectively, we investigated different strategies to transfer knowledge accumulated by BO to the EA in the form of an initial population. The best strategy that balances quality and diversity well is to cluster the top 50\% of all BO-generated solutions and select the best of each cluster to seed the EA. Additionally, we identified a heuristic value of the switch point to switch from BO to the EA in BEA. Based on preliminary experiments, we found that a value around 250-300 iterations (function evaluations) can be recommended. Furthermore, we also adjusted the EA part of BEA by a new self-adaptive Gaussian mutation, where the mutation step-sizes are also influenced by the actually measured time efficiency (gain per second).  

To assess BEA, we compared it to an EA and BO on three well-known benchmark objective functions with many local optima and nine test cases in evolutionary robotics. The results on the objective functions show that BEA clearly outperforms both the EA and BO not only according to the objective value, but also in terms of time efficiency. The robotic test cases confirmed these findings not only in simulations, but also in the real world. To this end, we installed the best robot controllers learned by BO and BEA on real robots and found that the BEA-optimized behaviors were better in almost all cases. Our future work aims at applying BEA to efficiently learn controllers in real time on real robots with various evolved morphologies. 


%



\section*{Acknowledgment}
The authors would like to thank Fuda van Diggelen for help with real robots and insightful remarks.

\ifCLASSOPTIONcaptionsoff
  \newpage
\fi

\bibliographystyle{IEEEtran}
\bibliography{boea19}

\end{document}